\documentclass[sigconf]{acmart}

\usepackage{balance}
\renewcommand\footnotetextcopyrightpermission[1]{} 
\AtBeginDocument{%
  \providecommand\BibTeX{{%
    \normalfont B\kern-0.5em{\scshape i\kern-0.25em b}\kern-0.8em\TeX}}}

\usepackage{bm} 
\usepackage{amsmath}
\usepackage{multirow}
\usepackage{bbding}

\usepackage{array}
\newcommand{\PreserveBackslash}[1]{\let\temp=\\#1\let\\=\temp}
\newcolumntype{C}[1]{>{\PreserveBackslash\centering}p{#1}}
\newcolumntype{R}[1]{>{\PreserveBackslash\raggedleft}p{#1}}
\newcolumntype{L}[1]{>{\PreserveBackslash\raggedright}p{#1}}

\renewcommand{\shortauthors}{Anonymous Author, et al.}

\begin{document}
\fancyhead{}

\title{Spatio-Temporal Transformer for Dynamic Facial Expression Recognition in the Wild}





\author{Fuyan Ma}
\affiliation{%
 \institution{College of Electrical and Information Engineering, \\Hunan University}
 \institution{Key Laboratory of Visual Perception and Artificial Intelligence of Hunan Province}
 \city{Changsha}
 \country{China}}
\email{mafuyan@hnu.edu.cn}

\author{Bin Sun}
\affiliation{%
 \institution{College of Electrical and Information Engineering, \\Hunan University}
 \institution{Key Laboratory of Visual Perception and Artificial Intelligence of Hunan Province}
 \city{Changsha}
 \country{China}}
\email{sunbin611@hnu.edu.cn}

\author{Shutao Li}
\affiliation{%
 \institution{College of Electrical and Information Engineering, \\Hunan University}
 \institution{Key Laboratory of Visual Perception and Artificial Intelligence of Hunan Province}
 \city{Changsha}
 \country{China}}
\email{shutao\_li@hnu.edu.cn}


\begin{abstract}
Previous methods for dynamic facial expression in the wild are mainly based on 
Convolutional Neural Networks (CNNs), whose local operations ignore the long-range dependencies in videos.
To solve this problem, we propose the spatio-temporal Transformer (STT) to capture discriminative features within each frame and model contextual relationships among frames.
Spatio-temporal dependencies are captured and integrated by our unified Transformer.
Specifically, given an image sequence consisting of multiple frames as input, we utilize the CNN backbone to translate each frame into a visual feature sequence.
Subsequently, the spatial attention and the temporal attention within each block are jointly applied for learning spatio-temporal representations at the sequence level.
In addition, we propose the compact softmax cross entropy loss to further encourage the learned features have the minimum intra-class distance and the maximum inter-class distance.
Experiments on two in-the-wild dynamic facial expression datasets (i.e., DFEW and AFEW) indicate that our method provides an effective way to make use of the spatial and temporal dependencies for dynamic facial expression recognition. The source code and the training logs will be made publicly available.
\end{abstract}




\maketitle

\section{Introduction}
Facial expression is one of the most important ways for human to convey their emotions and communicate with each other\cite{darwin2015expression}.
Facial expression recognition (FER) has been an emerging topic, due to its essential real-world applications in driver safety monitoring \cite{wilhelm2019towards}, human robot emotional interaction \cite{liu2017facial}, elderly healthcare\cite{bisogni2022impact}  and so on.
Accurately and automatically recognizing facial expression in videos is an interdisciplinary and integrated research, which spans from psychology to computer science.
Previous intensive studies (such as \cite{li2017reliable,zhao2021learning,ma2021facial}) have been conducted on static facial expression recognition (SFER). However, a facial expression is a dynamic process, which consists of various facial muscle motions in different facial regions.
Temporal facial expression cues are of significance for understanding the emotional state.

\begin{figure}[t]
	\centering
	\includegraphics[width=\linewidth]{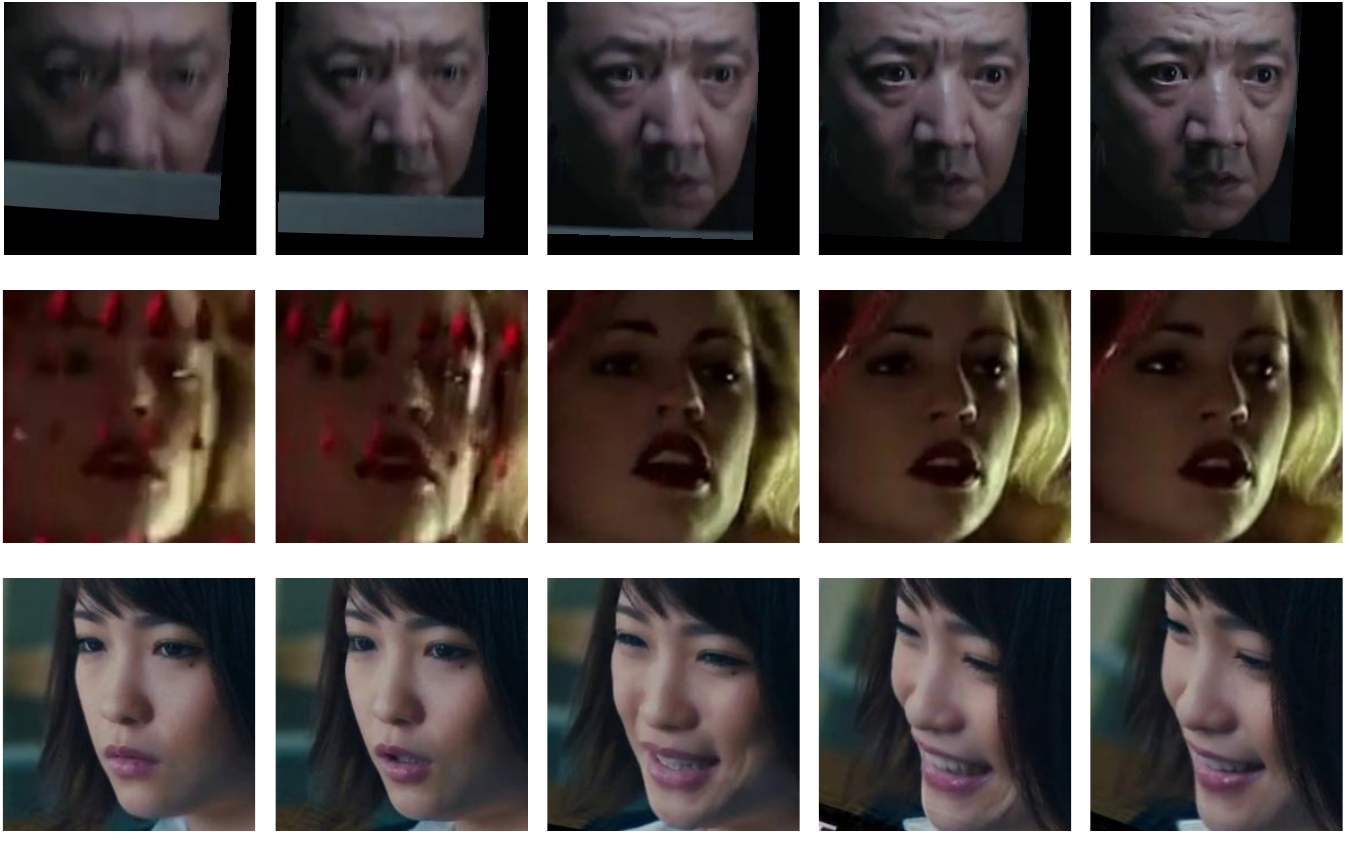}
	\caption{Samples from the DFEW \cite{jiang2020dfew} dataset. Variant head poses, poor illumination and occlusions can be seen from these image sequences.}
	\label{dfew_intro}
  \end{figure}

Researchers have focused more and more attention on dynamic facial expression recognition (DFER) in recent years.
DFER aims to classify a facial image sequence into several discrete expressions (such as happiness, surprise, neutral, sadness, fear, disgust, anger).
Constrained by the limited dynamic facial expression datasets, most of DFER methods are designed based on several lab-collected datasets, such as Oulu-CASIA\cite{zhao2011facial}, MMI\cite{valstar2010induced} and CK+\cite{lucey2010extended}.
The relatively few samples of these lab-collected datasets are not enough for training deep learning-based methods. 
The lack of large-scale databases seriously hinders the progress of dynamic facial expression recognition. Fortunately, Jiang \textit{et al.} \cite{jiang2020dfew} propose a novel and well-annotated dataset named DFEW, with samples shown in Fig. \ref{dfew_intro}.
 DFEW is the largest in-the-wild dynamic facial expression dataset with respect to the data volume and the sample sources, compared with other datasets (such as Aff-Wild \cite{kollias2019deep}, AFEW \cite{dhall2019emotiw} and CAER \cite{lee2019context}), which will definitely push forward the development of in-the-wild DFER.

Previous DFER methods can be mainly divided into two categories (i.e., static frame-based methods and dynamic sequence based methods) \cite{li2020deep}.
Most of the static frame-based methods utilize local binary patterns \cite{huang2010new}, Gabor wavelets \cite{lee2016collaborative}  and convolutional features\cite{yang2018facial,liu2018conditional} to select peak (apex) frames in videos, and further conduct facial expression recognition on these frames.
For example, Zhao \textit{et al.} \cite{zhao2016peak} propose to use a sample with peak expression to supervise the peak-piloted deep network to learn from a sample of non-peak expression. In addition, Meng \textit{et al.} \cite{meng2019frame} propose to use the attention mechanism to aggregate some discriminative frames into a single video representation, achieving superior performance on AFEW and CK+.
Although these methods perform well by selecting peak frames, they neglect the temporal dynamics and correlation among facial frames.

Different from static frame-based methods, dynamic sequence based methods usually use 3D convolution neural networks (3DCNN) \cite{ayral2021temporal}, long-short term memory (LSTM) \cite{vielzeuf2017temporal} to learn the spatio-temporal relationships, which can model long-term dependencies and improve the performance of DFER.
For instance, Kim \textit{et al.} \cite{kim2017multi} propose to use a LSTM network to learn the temporal characteristics of the spatial features, which achieves higher recognition rates.
Chen \textit{et al.} \cite{chen2020stcam} propose a 3D-Inception-Resnet to directly make the learned features more representative by calculating a spatial-temporal-wise and a channel-wise attention map.
Very recently, Li \textit{et al.} \cite{liu2022clip} utilize a CNN to extract clip-level features and re-weight each clip-based representation to achieve clip-aware dynamic facial expression.
Although several approaches have been proposed for in-the-wild DFER, the performances of these methods are still far from being satisfactory, because of occlusions, variant head poses, poor illumination and other unexpected issues in real-world scenes.

It is a challenging task to capture discriminative features in the spatial and temporal domain for in-the-wild DFER.
Recent flourishing of Transformer-based methods on computer vision tasks has considerably deepened our understanding about discriminative feature representation and contextual information modeling.
Therefore, we propose a simple but effective spatio-temporal Transformer (STT) for in-the-wild DFER, which can exploit and capture the facial appearance information in the spatial domain as well as the evolution information in the temporal domain to enhance the recognition performances.
Specifically, we firstly translate each facial frame from a video into a visual feature sequence.
Each patch of the feature map can be regard as a token of the feature sequence.
We elaborately design the spatio-temporal Transformer for capture discriminative feature tokens and model temporal dependencies among different frames.
To further increase the model discriminant ability, we impose the constraint on the prediction distribution by the loss function, following \cite{jiang2020dfew, kobayashi2019large, wen2016discriminative,liu2016large}.
We propose the compact softmax cross entropy loss to decrease the intra-class distance and increase the inter-class distance.
The quantitative results and the visualization results demonstrate the effectiveness of our method for in-the-wild dynamic facial expression recognition. We summarize the contributions of our method as follows:
\begin{itemize}
	\item We design a simple but effective spatio-temporal Transformer to capture discriminative features and model temporal relationships among facial frames. 
	\item Towards enhancing the intra-class correlation and maximizing the inter-class distance, we propose the compact softmax cross entropy loss to supervise our model in a regularization manner.
	\item The extensive results and the visualization results on two in-the-wild DFER datasets demonstrate the superiority of our method.
\end{itemize}

\section{Related Work}
\label{related_work}
\subsection{Dynamic Facial Expression Recognition in the Wild}
With the development of deep learning, especially Convolution Neural Networks, researchers have made significant progress on dynamic facial expression recognition in the wild.
An intuitive way to extract the facial expression sequence is to learn the spatial and temporal representations separately.
Under this paradigm, several methods \cite{kim2017multi,kuo2018compact,cai2016video} have been proposed to use a 2D CNN for extracting spatial features and apply a recurrent neural network (such LSTM) to model temporal correlation.
For example, Li \textit{et al.} \cite{liang2020deep} use a deep network to extract spatial features from each frame and model temporal dynamics by a convolutional network. The discriminative fused features are obtained by a Bi-LSTM network.
A global face module is proposed to learn spatial features from the peak frame and a LSTM module is used to extract temporal information \cite{yu2020facial}. Similarly, Zhang \textit{et al.} propose a novel framework named STRNN to integrate both spatial and temporal information separately. These methods use the spatio-temporal information of image sequence to improve the recognition performances.

Another way to learn the spatial and temporal representation is to model such correlation by the 3D CNN. Liu \textit{et al.} \cite{liu2014deeply} directly apply a 3D CNN with deformable facial action parts constraints to learn part-based representations for dynamic expression recognition.
Jung \textit{et al.} \cite{jung2015joint} also use the 3D CNN to learn spatial features along the time axis and aggregate temporal geometry features to boost the performance.
Although these methods employ the 3D CNN to capture the temporal correlation, they only use one or two layers with 3D convolutional filters and fail to model the complicated spatio-temporal correlation.
Very recently, Ayral \textit{et al.} \cite{ayral2021temporal} propose to re-weight different clips and achieve a clip-aware 3D CNN for dynamic facial expression, which outperforms other methods on AFEW.
A novel framework called EC-STFL \cite{jiang2020dfew} is also proposed to reduce the inter-class distance and enhance the intra-class correlation.
Although these deep-learning based methods can extract spatial and temporal information, they all decouple the spatial features and the temporal features at different stages.
We argue that if we can jointly get the spatial and temporal information, the performance of DFER will boost.

\begin{figure*}[t]
	\centering
	\includegraphics[width=\textwidth]{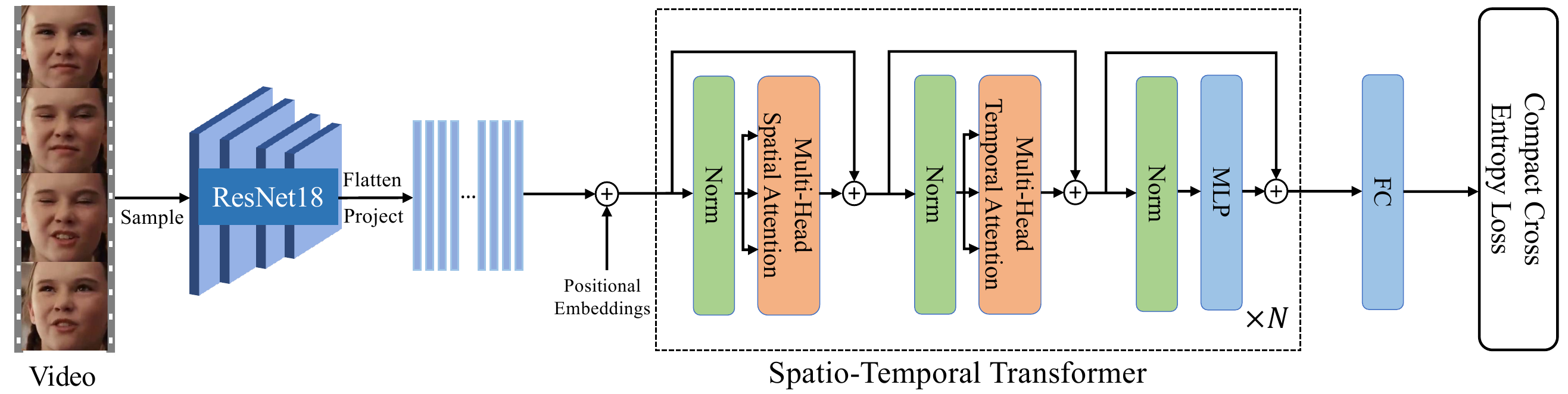}
	\caption{Overview of our method for dynamic facial expression recognition. The facial frames sampled from a video are processed by the CNN backbone to obtain frame-level feature sequences. The spatio-temporal Transformer jointly calculates the spatial attention and the temporal attention for capturing discriminative feature tokens.
	The compact cross entropy loss is utilized to further supervise our model to learn spatio-temporal representations, which have the close intra-class correlation and the large inter-class distance.}
	\label{STT}
  \end{figure*}

\subsection{Transformers for Facial Expression Recognition}
Transformer-based methods have shown dominant performance on various computer vision tasks, such as object detection \cite{beal2020toward}, image synthesis \cite{esser2021taming} and pose estimation \cite{yang2020transpose}.
ViT \cite{dosovitskiy2020image} directly apply a standard Transformer network for a facial patch sequence and prove its power on image classification.
Moreover, the Transformer-based methods \cite{bertasius2021space,sharir2021image} have been proposed for video understanding.
Regarding the facial expression recognition (FER) task, Ma \textit{et al.} \cite{ma2021facial} first use Transformers to dynamically modeling the relationships among facial patches, and achieve outstanding performances on occlusion and pose-variant FER.
Following this idea, Huang \textit{et al.} \cite{huang2021facial} use the visual Transformer to capture the relationships of different facial regions.
Gao \textit{et al.} also achieve occlusion-aware FER by a Transformer architecture.
Similarly, Li \textit{et al.} \cite{li2021mvt} propose a novel mask vision Transformer to focus on discriminative facial parts and filter out the complex backgrounds.
Very recently, Former-DFER\cite{zhao2021former} is proposed to separately extract spatial and temporal correlations by Transformers. Former-DFER achieves current state-of-the-art performances on DFER.
The self-attention mechanism of Transformers has the ability to capture long-range interactions among features tokens, which is of significance for the in-the-wild facial expression recognition.
Therefore, the applications of Transformers, especially for in-the-wild DFER, are worth exploring.

\section{Method}
\label{method}
\subsection{Overview}
The proposed framework is shown in Fig. \ref{STT}, which consists of three parts, i.e., input embedding generation, spatio-temporal Transformer and the compact softmax cross entropy loss.
We first use the uniform sampling strategy to obtain the fixed-length facial frame sequence from a video.
Then the CNN (i.e., ResNet18) is applied for extracting visual feature maps. We flatten and project the features maps as the input of the spatio-temporal Transformer (STT).
The STT utilizes the spatial attention and the temporal attention to jointly learn the discriminative features from the sequence.
Subsequently, the compact softmax cross entropy loss is used for minimizing the intra-class distance and maximizing the inter-class distance and further improving the recognition performance.

\subsection{Input Embedding Generation}

Given an image sequence $X \in \mathbb{R}^{F\times H_{0}\times W_{0}\times 3}$ with $F$ RGB facial frames of size $H_{0}\times W_{0}$ sampled from the video, we utilize the CNN backbone to extract frame-level features.
Specifically, we first divide the video into $S$ segments and uniformly sample $T$ consecutive frames from each segment for training. Finally, we obtain $F=S\times T$ training frames for each video sample. During inference, $T$ frames in the mid of each segment are heuristically selected, and the length of testing frames is also $F=S\times T$.

A standard CNN backbone (i.e., ResNet18) is used to generate high-level feature maps of size $H \times W$ for each frame. The clip-level features $\bm{f}_{0} \in \mathbb{R}^{F\times H \times W \times C}$ are obtained by concatenating all frame-level feature maps.
Then we flatten the spatial dimension of the clip-level feature maps $\bm{f}_{0}$ and project them by a $1 \times 1$ convolution, resulting in a new feature sequence $\bm{f}_{1} \in \mathbb{R}^{F\times (H\times W)\times d}$.
It is noted that the temporal order of $\bm{f}_{1}$ is in accordance with that of the input $X$.
To supplement the spatio-temporal positional information for the feature sequence $\bm{f}_{1}$, we incorporate the learnable positional embeddings with $\bm{f}_{1}$.
Specifically, we add the spatial positional embedding by
\begin{equation}
	\bm{f}_{2} = \bm{f}_{1} + e^{space}_{pos},
\end{equation}
where $e^{space}_{pos}\in \mathbb{R}^{1\times (H\times W)\times d}$ is a learnable spatial positional embedding.

We also prepend the classification token to the sequence $f_{2}$ at the temporal dimension, which models the global state of the sequence and is further used for recognition.
Then we reshape the new $\bm{f}_{2} \in \mathbb{R}^{(F+1)\times (H\times W)\times d}$ into $\bm{f}_{3} \in \mathbb{R}^{(H\times W) \times (F+1) \times d}$, and similarly add the temporal positional embedding by
\begin{equation}
	\bm{z}^{0} = \bm{f}_{3} + e^{time}_{pos},
\end{equation}
where $e^{time}_{pos} \in \mathbb{R}^{1 \times (F+1) \times d}$ is a learnable temporal positional embedding. Finally, the input embedding $\bm{z}^{0}$ to the spatio-temporal Transformer is obtained.

\subsection{Spatio-Temporal Transformer}
The spatio-temporal Transformer jointly utilizes the multi-head spatial attention and the multi-head temporal attention to learn a discriminative representation for each frame and exploit the long-range temporal dependencies to perform dynamic facial expression recognition. The spatio-temporal Transformer consists of $N$ blocks and each block is composed with the multi-head spatial attention and the multi-head temporal attention, which iteratively learns the contextual and discriminative spatio-temporal feature representation.
\par
{\bfseries Multi-Head Spatial Attention:}
At the $i$-th block of the spatio-temporal Transformer, we calculate the query/key/value vector for each patch of the same frame, which can be formatted by
\begin{equation}
	\label{eq3}
	\bm{q}^{(i,a)}_{(p,t)} = W^{(i,a)}_{Q}LN(\bm{z}^{l-1}_{(p,t)}),
\end{equation}
\begin{equation}
	\label{eq4}
	\bm{k}^{(i,a)}_{(p,t)} = W^{(i,a)}_{K}LN(\bm{z}^{l-1}_{(p,t)}),
\end{equation}
\begin{equation}
	\label{eq5}
	\bm{v}^{(i,a)}_{(p,t)} = W^{(i,a)}_{V}LN(\bm{z}^{l-1}_{(p,t)}),
\end{equation}
where $a$ is an index over multiple spatial heads, and $p \in \{1,\dots, H\times W\}$, $t \in \{1,\dots, F\}$ are the patch and frame index, respectively. In addition, $W_{Q}, W_{K}, W_{V}$ are the parameters of linear projections. The hidden dimensions of $\bm{q}$, $\bm{k}$, $\bm{v}$ for each head is set to  $D_{h} = d/N_{h}$, and $N_{h}$ is the number of multiple heads.

\begin{figure}[t]
	\centering
	\includegraphics[width=0.6\linewidth]{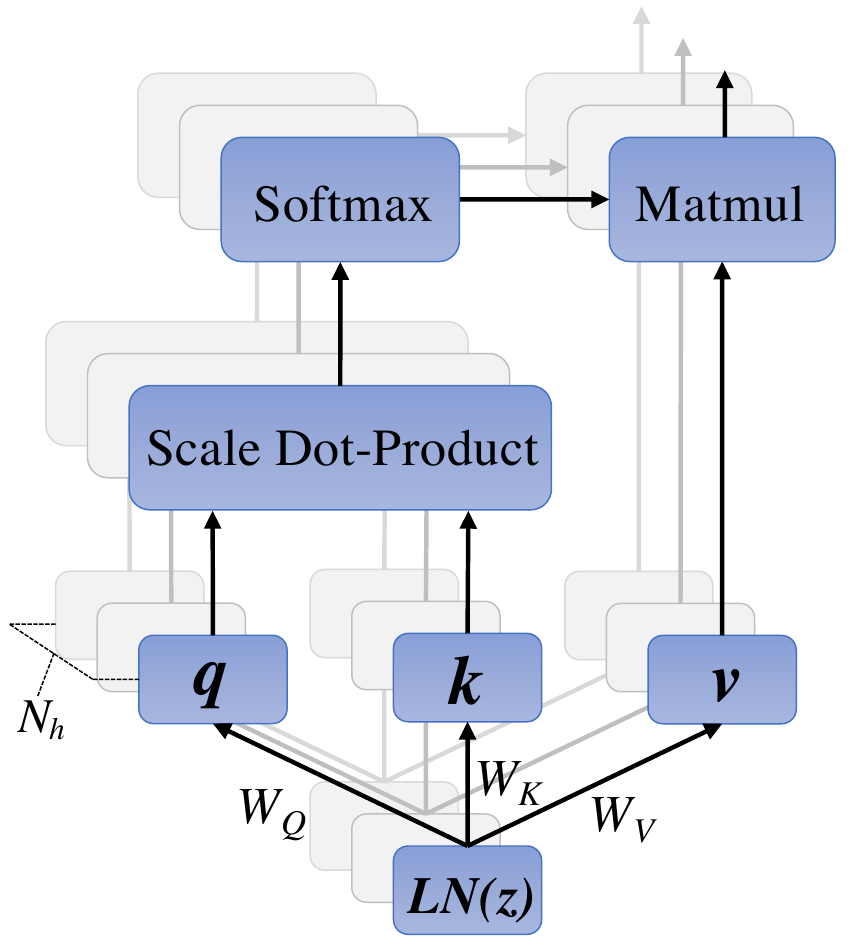}
	\caption{Illustration of the multi-head spatial attention.}
	\label{mhsa}
  \end{figure}

The spatial attention is applied between feature patches of the same frame.
Therefore, the spatial attention weights are obtained by the scale dot-product by:
\begin{equation}
	\bm{\alpha}^{(i,a) space}_{(p,t)} = SM\Bigg(\frac{{\bm{q}^{(i,a)}_{(p,t)}}^\mathrm{T}}{\sqrt{D_{h}} }  \cdot \bigg\{ \bm{k}^{(i,a)}_{(p',t)} \bigg\}_{p'=1, \dots, H\times W} \Bigg),
\end{equation}
where $SM$ represents the softmax activation function.
The output of each head is given by
\begin{equation}
	\bm{s}^{(i,a)}_{(p,t)} = \sum_{p'}^{H\times W}\bm{\alpha}^{(i,a) space}_{(p,t),(p',t)}\bm{v}^{(i,a)}_{(p',t)}
\end{equation}
Subsequently, the outputs are concatenated and projected, using the residual connection:
\begin{equation}
	\bm{z'}^{(i)}_{(p,t)} = W'_{O}\bigg[\bm{s}^{(i,1)}_{(p,t)},\dots,\bm{s}^{(i,N_{h})}_{(p,t)} \bigg]^{\mathrm{T}} + \bm{z}^{(i-1)}_{(p,t)},
\end{equation}
where $W'_{O}$ denote the parameters of a linear projection for the concatenated multi-head features.
Although the multi-head spatial attention provides a powerful representation in the spatial dimension, it does not reflect the temporal dependencies across different facial frames.

{\bfseries Multi-Head Temporal Attention:}
To further capture the temporal correlation within the frame sequence, we apply the multi-head temporal attention on the encoding $\bm{z'}^{(i)}_{(p,t)}$ generated by the multi-head spatial attention.
Towards obtaining the multi-head temporal attention, we make $F+1$ query-key comparisons, using each patch $(p,t)$ with all patches at the same spatial location in the other frames.
The query/key/value vector $(\bm{q'}^{(i,a)}_{(p,t)}, \bm{k'}^{(i,a)}_{(p,t)}, \bm{v'}^{(i,a)}_{(p,t)})$ for each patch $(p,t)$ is generated similarly in Eq. \ref{eq3}, Eq. \ref{eq4}, Eq. \ref{eq5}, respectively.
Considering that a classification token is prepended to the frame feature sequence in the temporal dimension, we calculate the temporal attention weights by 
\begin{equation}
	\bm{\alpha'}^{(i,a) time}_{(p,t)} = SM\Bigg(\frac{{\bm{q'}^{(i,a)}_{(p,t)}}^\mathrm{T}}{\sqrt{D_{h}} }  \cdot \Bigg[ \bm{k'}^{(i,a)}_{(0,0)} \cdot \bigg \{ \bm{k'}^{(i,a)}_{(p,t')} \bigg \}_{t'=1, \dots, F} \Bigg] \Bigg),
\end{equation}
where $\bm{k'}^{(i,a)}_{(p,0)}$ is the corresponding key value of the classification token at frame $0$.
The temporal attention weights $\bm{\alpha'}^{(i,a) time}_{(p,t)}$ are used to sum over the values for each temporal head:
\begin{equation}
	\bm{s'}^{(i,a)}_{(p,t)} = \bm{\alpha'}^{(i,a) time}_{(p,0)}\bm{v'}^{(i,a)}_{(p,0)} + \sum_{t'=1}^{F} \bm{\alpha'}^{(i,a) time}_{(p,t')}\bm{v'}^{(i,a)}_{(p,t')}.
\end{equation}
The outputs from all temporal heads are concatenated, projected, and passed through a multi-layer perception with a GELU \cite{hendrycks2016gaussian} activation function, which can be defined as:
\begin{equation}
	\bm{z''}^{(i)}_{(p,t)} = W''_{O}\bigg[\bm{s'}^{(i,1)}_{(p,t)},\dots,\bm{s'}^{(i,N_{h})}_{(p,t)} \bigg]^{\mathrm{T}} + \bm{z'}^{(i)}_{(p,t)},
\end{equation}

\begin{equation}
	\bm{z}^{(i)}_{(p,t)} = MLP(LN(\bm{z''}^{(i)}_{(p,t)})) + \bm{z''}^{(i)}_{(p,t)}.
\end{equation}
The spatio-temporal encoding $\bm{z}^{(i)}_{(p,t)}$ of the $i$-th block serves as the input to the $(i+1)$-th block.
And we apply a single fully connected (FC) layer to the classification token $\bm{z}^{(N)}_{(0,0)}$ of the final block:
\begin{equation}
	\bm{p} = FC(\bm{z}^{(N)}_{(0,0)}),
\end{equation}
where $p$ is the prediction distribution of $C$ facial expression classes.

\subsection{Compact Softmax Cross Entropy Loss}
The cross softmax entropy loss with softmax is undoubtedly the most common used supervision loss for the facial expression recognition task.
However, learning discriminative spatio-temporal features for in-the-wild DFER requires the loss function to have the ability of maximizing the feature distance between different categories. Therefore, we need a more versatile loss function for DFER.

The label smoothing strategy has been a great success in training deep CNNs, which helps the models overcome the overfitting issue and learn more discriminative features.
Given a training sample $(x,y)$ and the prediction $\hat{y}$, the label smoothing strategy performs label regularization by introducing uniform distribution $\bm{u}$ over $C$ classes on the ground-truth label distribution $\bm{q}$:
\begin{equation}
	\bm{s} = \bm{q} + \lambda \bm{u},
\end{equation}
where $\lambda$ is the balancing factor and $\bm{q}$ can be denoted as $\bm{q}(\hat{y} =y|x)=1$ and $\bm{q}(\hat{y} \neq y|x)=0$.
Therefore, we compute the cross entropy loss by
\begin{equation}
	\mathcal{L}_{soft} = -\sum_{k=1}^{C}\bm{s}_{k}\log(\bm{p}_{k}),
\end{equation}
where $\bm{s}_k$, $\bm{p}_{k}$ are the soft label and the predicted probability for the target category $k$, respectively.
Furthermore, the loss $\mathcal{L}_{soft}$ can be rewritten as:
\begin{equation}
	\begin{aligned}
		\mathcal{L}_{soft} &= -(1-\lambda)\mathcal{L}_{ce} - \lambda \mathcal{L}_{reg}\\
		 &= -(1-\lambda)\log(\bm{p}_y) -  \frac{\lambda}{C}\sum_{c=1}^C \log(\bm{p}_c) \\
		 &= -(1-\lambda)\log(\bm{p}_y) + \lambda\{\mathcal{D}(\bm{u}||\bm{p}) + \log(C) \}   , \\
	\end{aligned}
\end{equation}
where ${L}_{ce}$ and $\mathcal{L}_{reg}$ denotes the standard cross entropy loss and a regularization term. 
The Kullback-Leibler (KL) divergence $\mathcal{D}(\bm{u}||\bm{p})$ measures the difference between the two distributions $\bm{u}$ and $\bm{p}$, and suppresses the diversity of $\bm{p}$ in the loss $\mathcal{L}_{soft}$.

Based on the label smoothing regularization, we design the compact softmax cross entropy loss to decrease the intra-class distance and increase the inter-class distance.
Specifically, we leverage the symmetric KL divergence as the regularization term and impose the constraint on the prediction distribution $\bm{p'}$, with the target prediction excluded. By combining with the standard softmax cross entropy loss, our loss results in
\begin{equation}
	\begin{aligned}
	\mathcal{L}_{compact} = \mathcal{L}_{ce} &+ \beta \mathcal{L'}_{reg}\\
	=\log(\bm{p}_y) +  \frac{\beta}{2}\{\mathcal{D}(\bm{u'}||\bm{p'})\ &+ \mathcal{D}(\bm{p'} ||\bm{u'})\},\\
	\end{aligned}
\end{equation}
of which the KL divergences are given by 
\begin{equation}
	\mathcal{D}(\bm{u'}||\bm{p'}) = \sum_{c \neq y}\frac{1}{C-1}\log\big(\frac{1}{(C-1)\bm{p'}}\big),
\end{equation}
\begin{equation}
	\mathcal{D}(\bm{p'}||\bm{u'}) = \sum_{c \neq y}\log\big(\frac{1}{(C-1)\bm{p'}}\big).
\end{equation}
The uniform distribution $\bm{u'}$ is generated over $C-1$ classes, excluding the target $y$. Additionally, we calculate the new predicted distribution $\bm{p'}$ using the softmax function:
\begin{equation}
	\bm{p'} = softmax(\{\hat{y}_c^{logits}\}_{c\neq y}),
\end{equation}
where $\{\hat{y}_c^{logits}\}_{c\neq y}$ represents the non-target predicted logits.
With the regularization term $\mathcal{L'}_{reg}$, our loss function guides the model to increase the uniformity of the non-target predicted logits.
To summarize, our proposed loss function has the ability to induce the smaller intra-class distance and the larger inter-class distance, which is crucial to in-the-wild DFER.

\section{Experiments}
\label{experiments}
\subsection{Datasets}

{\bfseries DFEW} \cite{jiang2020dfew} is the largest publicly available DFER dataset, with 16,372 in-the-wild facial expression video clips.
All video clips are extracted from more than 1500 movies, whose themes cover tragedy, comedy, love and so on.
Due to the unconstrained conditions, DFEW is a significantly challenging dataset, of which the samples suffer occlusions, head pose changes, extreme illuminations and other problems.
Each clip of DFEW is labelled by ten independent professionally trained annotators with one of seven basic facial expressions (i.e., happiness, surprise, neutral, sadness, fear, disgust, anger), indicating that DFEW is a reliable dataset with very few annotation errors.
Similar with other FER databases, DFEW has the imbalanced category distribution issue.
To better evaluate different methods on the selected 12,509 clips, five-fold cross-validation protocol is adopted.

{\bfseries AFEW} has served as an evaluation dataset for the annual EmotiW from 2013 to 2019.
The videos of AFEW are close to real-world data with a wide range of challenges (e.g., complicated backgrounds, different head poses and movements), because the video samples are collected from TV serials and movies.
Same with DFEW, each video in AFEW is assigned to seven emotion labels.
AFEW consists of 773 videos from 67 movies for training, 383 samples from 33 movies for validation.
Since the official test dataset is unavailable, we report the results on the validation dataset to make a fair comparison with other state-of-the-art methods.

\subsection{Implementation Details}

{\bfseries Data Preprocessing:}
For DFEW, the processed face region images are publicly available, officially detected and aligned by Megvii Face API \cite{face++} and SeetaFace \cite{liu2017viplfacenet}.
We directly use the processed data for fair comparison.
For AFEW, the FFmpeg toolkit \cite[]{tomar2006converting} is used to extract frames from the raw videos. 
We then detect and align faces from the frames by the Dlib toolbox \cite{dlib09}.
All the face region images are resized to 112$\times$112 pixels as the inputs to our model.

{\bfseries Training and Testing Settings:}
We train our model with the PyTorch \cite{paszke2019pytorch} platform with two NVIDIA GTX 1080Ti GPU cards.
Our model is trained with the batch-size of 32 on DFEW and AFEW for 100 epochs and 20 epochs, respectively.
The SGD optimizer \cite{ruder2016overview} with an initial learning rate of 0.01 is used to optimize our proposed model on DFEW. 
The learning rate is divided by 10 every 40 epochs during training.
Since AFEW is a relatively tiny dataset, we follow Former-DFER \cite{zhao2021former} to fine-tune the model on AFEW by the pretrained weights on DFEW (fd1). The learning rate on AFEW is initialized to 0.001.
We use the pretrained ResNet18 on MS-Celeb-1M \cite{guo2016ms} as our CNN backbone.
During training, each video clip is divided into $S=8$ segments and uniformity sample $T=2$ consecutive frames from each segment.
During testing, $S$ and $T$ keep the same settings, and $T=2$ samples are selected from the mid of each segment, result in 16 facial frames as the input.
The number of spatio-temporal Transformer layers $N$ and the number of heads $a$ are empirically assigned to 4 and 8, respectively.
In addition, the hidden dimension $d$ is set to 512 and the loss balancing factor $\beta$ is set to 0.2.
The unweighted average recall (UAR) and the weighted average recall (WAR) serves as the evaluation metrics.

\subsection{Ablation Study}
We conduct the ablation experiments to analyze the impact of each component of our model (i.e., spatio-temporal Transformer and the compact softmax cross entropy loss).
All the experiments are conducted on the DFEW dataset with 5-fold cross validation.

\begin{figure*}[t]
	\centering
	\includegraphics[width=\linewidth]{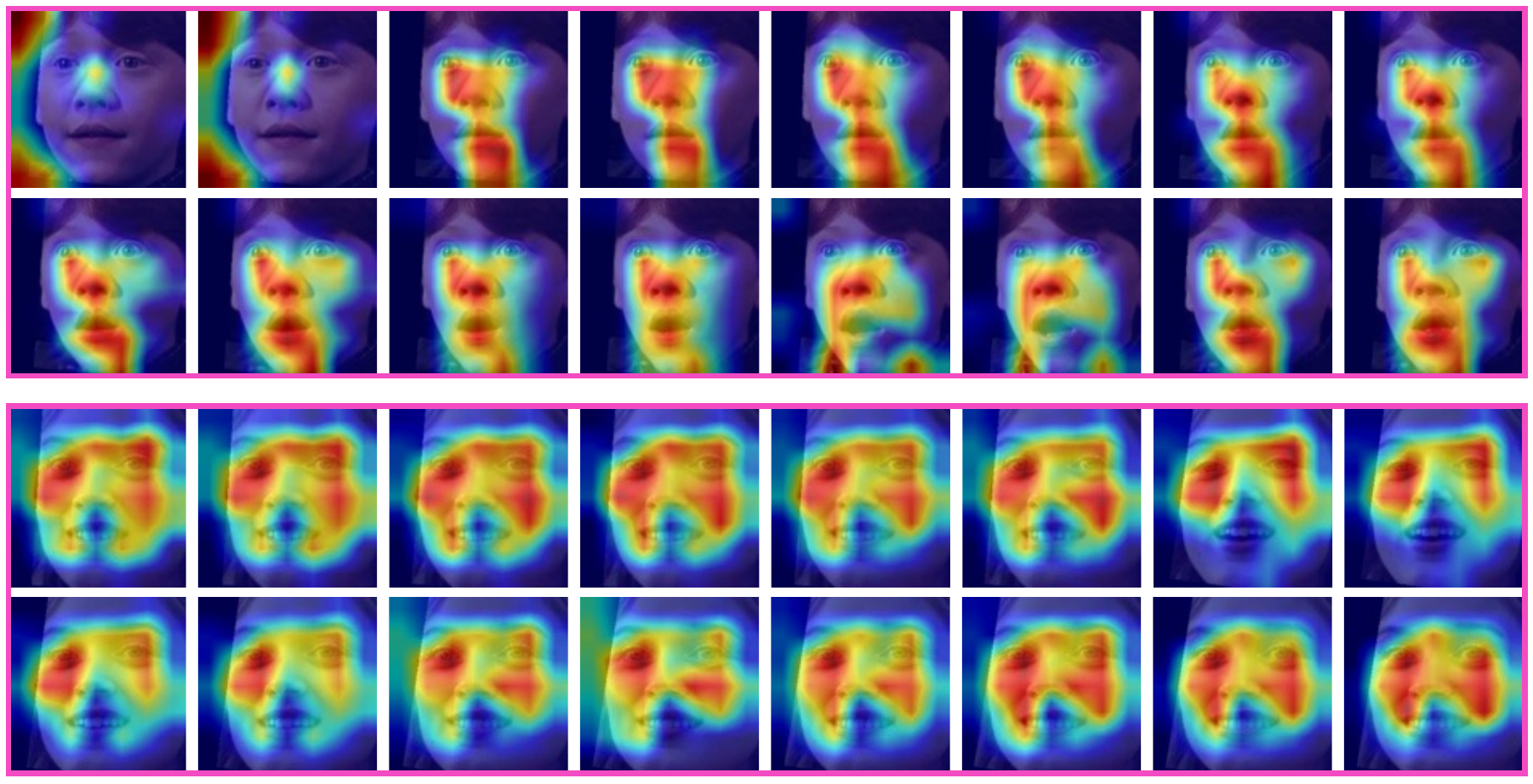}
	\caption{Visualization of the activation maps generated by Grad-CAM \cite{selvaraju2017grad}. Two frame sequences are presented to demonstrate the discriminative facial regions and the temporal correlations learned by our model.}
	\Description{}
	\label{cam}
  \end{figure*}

{\bfseries Importance of the spatio-temporal Transformer:} We first explore the impact of multi-head spatial attention and multi-head spatial attention.
The experimental results on DFEW in terms of the UAR and the WAR are shown in Tab. \ref{ab1}.
Our proposed spatio-temporal Transformer applies the temporal attention after the spatial attention in one block.
To demonstrate things their influences, we replace the spatio-temporal Transformer with a simple LSTM as the baseline.
The baseline achieves the UAR and the WAR at 52.02\% and 64.48\% respectively.
The use of multi-head spatial attention improves the UAR and the WAR by 1.13\% and 0.75\% over the baseline.
And employing the multi-head temporal attention improves the baseline by 1.28\% and 0.91\% in terms of the UAR and the WAR.
Moreover, when the spatial attention and the temporal attention are both applied, the UAR and the WAR are enhanced significantly by 2.56\% and 2.71\% over the baseline, respectively.

\begin{table}[t]
	\caption{Ablation study w.r.t. multi-head spatial attention, multi-head temporal attention, performed on DFEW with 5-fold cross validation.}
	\centering
	\begin{tabular}{cc|cc}
	\toprule
	\multirow{2}{*}{Multi-Head Spatial } & \multirow{2}{*}{Multi-head Temporal } & \multicolumn{2}{c}{Metrics (\%)} \\
	\cmidrule{3-4}        Attention 				       &      Attention            & UAR             & WAR            \\
	\midrule
   \XSolidBrush				           &    \XSolidBrush                     &     52.02           &         64.48       \\
   \CheckmarkBold					   &    \XSolidBrush                     &     53.15            &        65.23        \\
   \XSolidBrush			    		   &    \CheckmarkBold                   &     53.30            &        65.39        \\
   \CheckmarkBold					   &    \CheckmarkBold                &        54.58       &         66.65         \\
   \bottomrule
	\end{tabular}
	\label{ab1}
\end{table}

To further demonstrate the impact of the spatial attention and the temporal attention, we visualize the activation maps by Grad-CAM \cite{selvaraju2017grad} in Fig. \ref{cam}.
As shown in the first frame sequence of Fig. \ref{cam}, we can find that the first two facial frames are not highlighted on the facial regions. We speculate that our model can learn the temporal correlations among the frames, because the first two frames contribute very less to the target category surprise.
Our model can concatenate on the facial regions of the other frames of the first sequence, which indicates the effectiveness of our spatio-temporal Transformer for in-the-wild DFER.
The activation maps in the frames from the second sequence also support our conclusion.

\begin{figure*}[t]
	\centering
	\includegraphics[width=\linewidth]{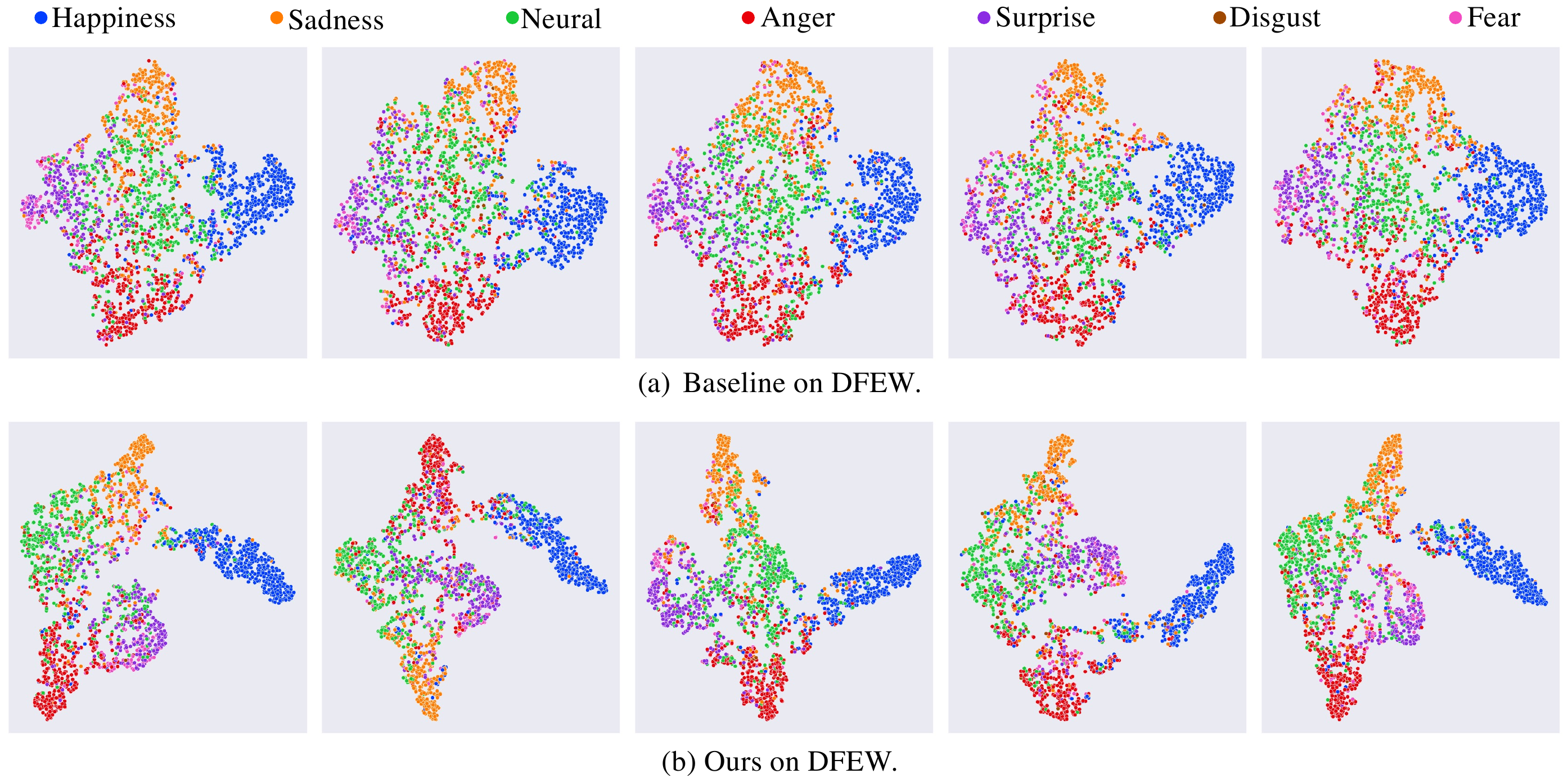}
	\caption{ Visualization of the feature distribution generated by t-SNE \cite{van2008visualizing} on DFEW (fd1$\sim$fd5). (a) denotes our model with the standard cross entropy loss. (b) represents our model with the compact cross entropy loss.}
	\Description{}
	\label{tsne}
	\end{figure*}

{\bfseries Effect of the compact softmax cross entropy loss:}
We then investigate the influences of different loss functions (i.e., the standard cross entropy loss, label smoothing, our compact loss). The corresponding results are shown in Tab. \ref{loss}.
With the standard cross entropy loss, our model still achieves the UAR of 54.15\% and the WAR of 66.39\%.
Label smoothing and our compact softmax cross entropy loss both improve the performance of our model purely with the cross entropy loss.
Specifically, our model trained with label smoothing achieves the gains of 0.14\% and 0.11\% in terms of the UAR and the WAR.
Our compact softmax cross entropy loss achieves the UAR of 54.58\% and the WAR of 66.65\%, which are 0.43\% and 0.26\% better than that with the standard softmax cross entropy loss.

\begin{table}[t]
	\caption{Evaluation of different loss functions on DFEW in terms of the UAR and the WAR.}
	\centering
	\begin{tabular}{c|cc}
	\toprule
	 \multirow{2}{*}{Loss Function } & \multicolumn{2}{c}{Metrics (\%)} \\
	\cmidrule{2-3}                     & UAR             & WAR            \\
	\midrule
	Cross Entropy Loss                     & 54.15                &     66.39           \\
	Label Smoothing                     &   54.29              &    66.50            \\
	Ours                   &   54.58              &        66.65        \\
   \bottomrule
	\end{tabular}
	\label{loss}
\end{table}

\begin{figure}[t]
	\centering
	\includegraphics[width=\linewidth]{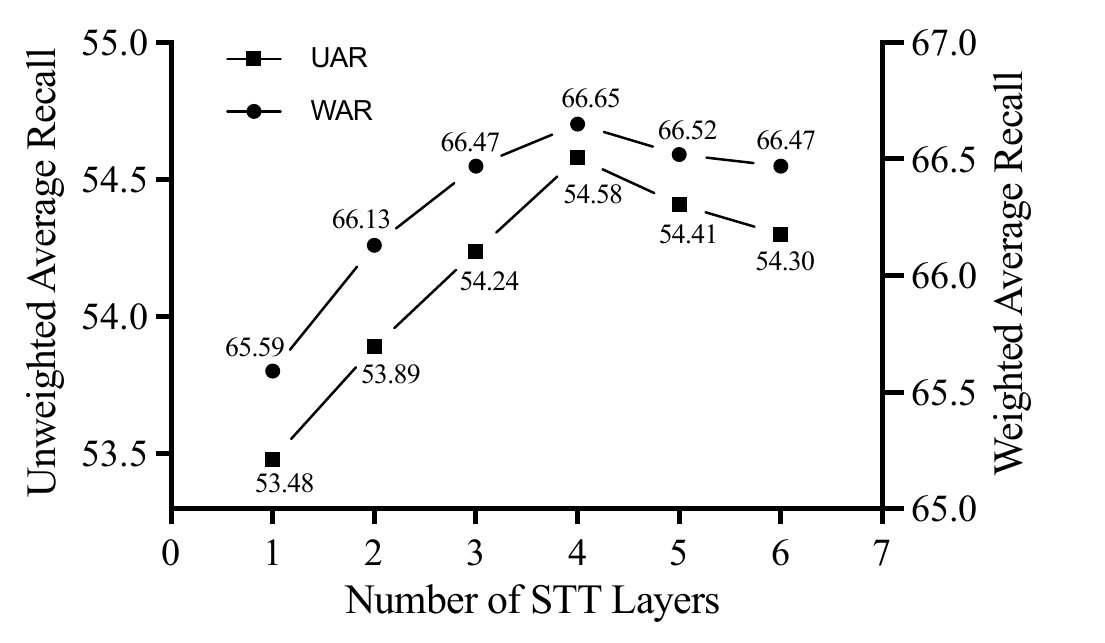}
	\caption{Evaluation of our method with different layers on DFEW.}
	\Description{}
	\label{ab3}
	\end{figure}

To better understand the effect of our proposed loss, we utilize t-SNE \cite{van2008visualizing} to visualize the learned features on DFEW (fd1$\sim$fd5) in Fig. \ref{tsne}.
The feature distribution on a 2D plane can reflect the intra-class compactness and the inter-class distance.
As shown in Fig. \ref{tsne}, our compact loss enables the learned features have a better aggregation effect and show more clear inter-class boundaries among different expressions, compared with the standard cross entropy loss.
Such patterns in Fig. \ref{tsne} also prove that our model has the ability to learn more discriminative features under the guidance of our proposed loss function.

{\bfseries Impact of Hyper-parameters:}
Finally, we study the impact of different number of spatio-temporal Transformer layers.
We vary the number of layers $N \in \{1,2,3,4,5,6\}$ to conduct experiments.
$N$ is a trade-off parameter for balancing the recognition performance and the complexity of the model.
The experimental results are shown in Fig. \ref{ab3}.
It is apparent that choosing the value of $N=4$ obtains the best performances of 54.58\% and 66.65\% in terms of the UAR and the WAR, respectively.
The results also indicates that the shallower model's feature representation ability is more limited.

\subsection{Comparison with State-of-the Art Methods}
We compare our method with other methods on DFEW (5-fold cross validation) and AFEW, with respect to the UAR, the WAR.

\begin{table*}[t]
	\caption{Comparison with other state-of-the-art methods on DFEW with 5-fold cross validation. The best results are in bold. \underline{Underline} represents the second best. TI denotes time interpolation and DS denotes dynamic sampling. The evaluation metrics include the unweighted average recall (UAR) and the weighted average recall (WAR).}
	\begin{tabular}{c|c|ccccccc|cc}
		\toprule

	\multirow{2}{*}{Method} & Sample & \multicolumn{7}{c|}{Accuracy of Each Emotion (\%)}  & \multicolumn{2}{c}{Metrics (\%)} \\
	\cmidrule{3-11}
	&  Strategies     & Happiness & Sadness & Neutral & Anger & Surprise & Disgust & Fear  & UAR  & WAR    \\
	\midrule
	C3D \cite{tran2015learning}                     & TI                                                                           & 75.17     & 39.49   & 55.11   & 62.49 & 45.00    & 1.38    & 20.51 & 42.74           & 53.54           \\
	P3D \cite{qiu2017learning}                     & TI                                                                           & 74.85     & 43.40   & 54.18   & 60.42 & 50.99    & 0.69    & 23.28 & 43.97           & 54.47            \\
	R(2+1)D18 \cite{tran2018closer}               & TI                                                                           & 79.67     & 39.07   & 57.66   & 50.39 & 48.26    & 3.45    & 21.06 & 42.79           & 53.22           \\
	3D Resnet18 \cite{hara2018can}            & TI                                                                           & 73.13     & 48.26   & 50.51   & 64.75 & 50.10    & 0.00    & 26.39 & 44.73           & 54.98           \\
	I3D-RGB \cite{carreira2017quo}                & TI                                                                           & 78.61     & 44.19   & 56.69   & 55.87 & 45.88    & 2.07    & 20.51 & 43.40           & 54.27           \\
	VGG11+LSTM  \cite{simonyan2014very, hochreiter1997long}            & TI                                                                           & 76.89     & 37.65   & 58.04   & 60.70 & 43.70    & 0.00    & 19.73 & 42.39           & 53.70           \\
	ResNet18+LSTM \cite{he2016deep, hochreiter1997long}          & TI                                                                           & 78.00     & 40.65   & 53.77   & 56.83 & 45.00    & \textbf{4.14}    & 21.62 & 42.86           & 53.08            \\
	3D R.18+Center Loss \cite{hara2018can,wen2016discriminative}     & TI                                                                           & 78.49     & 44.30   & 54.89   & 58.40 & 52.35    & 0.69    & 25.28 & 44.91           & 55.48             \\
	EC-STFL \cite{jiang2020dfew}                & TI                                                                           & 79.18     & 49.05   & 57.85   & 60.98 & 46.15    & 2.76    & 21.51 & 45.35           & 56.51            \\
	3D Resnet18 \cite{hara2018can}            & DS                                                                           & 76.32     & 50.21   & 64.18   & 62.85 & 47.52    & 0.00    & 24.56 & 46.52           & 58.27           \\
	ResNet18+LSTM  \cite{he2016deep, hochreiter1997long}         & DS                                                                           & 83.56     & 61.56   & \textbf{68.27}   & 65.29 & 51.26    & 0.00    & 29.34 & 51.32           & 63.85             \\
	Resnet18+GRU  \cite{he2016deep, chung2014empirical}          & DS                                                                           & 82.87     & 63.83   & 65.06   & 68.51 & 52.00    & 0.86    & 30.14 & 51.68           & 64.02            \\
	Former-DFER  \cite{zhao2021former}           & DS                                                                           & 84.05     & 62.57   & \underline{67.52}   & 70.03 & \textbf{56.43}    & 3.45    & 31.78 & 53.69           & 65.70               \\
	\midrule
	\textbf{STT (Ours)}              & DS    &  \textbf{87.36}         &   \textbf{67.90}      &  64.97       &  \textbf{71.24}     &  \underline{53.10}        & \underline{3.49}        &  \textbf{34.04}     &    \textbf{54.58}     &  \textbf{66.65}              \\
\bottomrule	
\end{tabular}
\label{dfew}
\end{table*}

\begin{table}[t]
	\caption{Comparison with other state-of-the-art methods on AFEW. The best results are in bold. }
	\begin{tabular}{c|c|cc}
		\toprule
		\multirow{2}{*}{Methods} & Sample     & \multicolumn{2}{c}{Metrics (\%)}  \\
		\cmidrule{3-4}
		& Strategies & UAR             & WAR            \\
	\midrule
	EmotiW-2019 Baseline \cite{dhall2019emotiw}     & -                                                                            & -               & 38.81                                                                                  \\
	C3D \cite{tran2015learning}         & DS                                                                           & 43.75           & 46.72                                                                              \\
	I3D-RGB \cite{carreira2017quo}                 & DS                                                                           & 41.86           & 45.41                                                                               \\
	R(2+1)D \cite{tran2018closer}                 & DS                                                                           & 42.89           & 46.19                                                                              \\
	3D ResNet18  \cite{hara2018can}            & DS                                                                           & 42.14           & 45.67                                                                                \\
	ResNet18+LSTM \cite{he2016deep, hochreiter1997long}           & DS                                                                           & 43.96           & 48.82                                                                               \\
	ResNet18+GRU \cite{he2016deep, chung2014empirical}            & DS                                                                           & 45.12           & 49.34                                                                               \\
	Former-DFER \cite{zhao2021former}             & DS                                                                           & 47.42           & 50.92                                                                                 \\
	\midrule
	\textbf{STT(ours)}       & DS                                                                           & \textbf{49.11}           & \textbf{54.23}                                                                      \\           
\bottomrule
\end{tabular}
\label{afew}
\end{table}

{\bfseries Comparison on DFEW:}
To make a fair comparison, we compare our methods with current state-of-the-art spatio-temporal networks on DFEW, which can be divided into 3D CNNs-based methods and 2D CNN-based methods.
The comparison results are shown in Tab. \ref{dfew}.
Our method obtains the best results using both metrics.
Specifically, Former-DFER is the previous state-of-the-art method with the UAR of 53.69\% and the WAR of 66.65\%.
Our method outperforms Former-DFER by 0.89\% and 0.95\% in terms of the UAR and the WAR, respectively.
Moreover, our method improves EC-STFL by 9.23\% and 10.14\% in UAR and WAR, which also aims to enhance the intra-class correlation and increase the inter-class distance.
As shown in Tab. \ref{dfew}, our method also obtains better results with respect to the category-level accuracy compared with other methods.
Specifically, our method achieves the highest accuracies on "happiness", "sadness", "anger" and "fear", and obtains the second best results on "surprise" and "disgust".
It is noted that the performances of the expressions "disgust" and "fear" are poor, which are mainly due to insufficient training samples.

\begin{figure}[t]
	\centering
	\includegraphics[width=\linewidth]{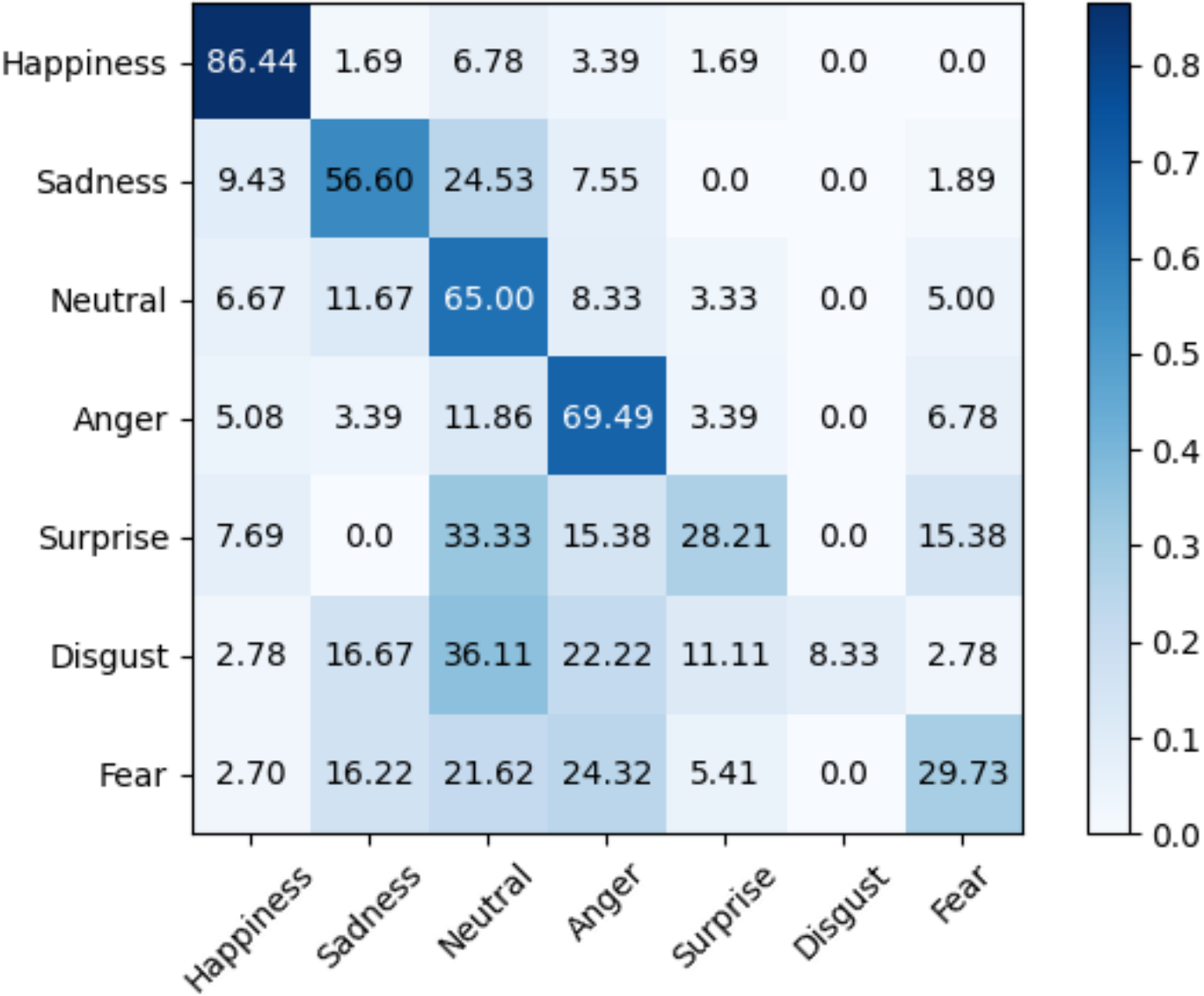}
	\caption{Experimental results of confusion matrix on AFEW.}
	\Description{}
	\label{con_afew}
\end{figure}

{\bfseries Comparison on AFEW:} We follow previous methods to conduct transfer learning experiments from DFEW to AFEW. 
The pretrained weights on DFEW (fd1) are used to initialize the model weights for AFEW.
The comparison results are shown in Tab. \ref{afew}.
Our method also achieves the best results in UAR and WAR compared with other state-of-the-art methods.
Specifically, the performances of our method are 1.69\% and 3.31\% better than these of Former-DFER.
In addition, our method improves the UAR and the WAR of 3D ResNet18 by 6.97\% and 8.56\%, respectively.
We provide the detailed recognition results of different facial expressions by the confusion matrix in Fig. \ref{con_afew}.
Similar with the results on DFEW, the performances of the expressions "disgust" and "fear" are still far from being satisfactory and can be improved in the future work.


\section{Conclusion}
\label{conclusion}
In this paper, we propose a simple but effective spatio-temporal Transformer and the compact softmax cross entropy loss for in-the-wild dynamic facial expression recognition (DFER).
Specifically, the spatio-temporal Transformer is utilized to handle DFER with the real-world scenario issues (e.g., occlusions, head poses and illuminations) by sequence modeling.
We jointly apply the spatial attention and the temporal attention within each block to learn spatio-temporal representations iteratively.
The spatial facial cues and the temporal dependencies are captured by a sequence of frame-level feature tokens from a global perspective.
To further increase the model discriminant ability, we impose the constraint on the prediction distribution by the compact loss to enhance the intra-class correlation and increase the inter-class distance.
The experimental results and the visualization results demonstrate that our method learns discriminative spatio-temporal feature representations and enhances the classification margin, which result in our favorable performance compared with other state-of-the-art methods.




\bibliographystyle{ACM-Reference-Format}

\bibliography{IEEEabrv,arxiv}


\begin{thebibliography}{63}


\ifx \showCODEN    \undefined \def \showCODEN     #1{\unskip}     \fi
\ifx \showDOI      \undefined \def \showDOI       #1{#1}\fi
\ifx \showISBNx    \undefined \def \showISBNx     #1{\unskip}     \fi
\ifx \showISBNxiii \undefined \def \showISBNxiii  #1{\unskip}     \fi
\ifx \showISSN     \undefined \def \showISSN      #1{\unskip}     \fi
\ifx \showLCCN     \undefined \def \showLCCN      #1{\unskip}     \fi
\ifx \shownote     \undefined \def \shownote      #1{#1}          \fi
\ifx \showarticletitle \undefined \def \showarticletitle #1{#1}   \fi
\ifx \showURL      \undefined \def \showURL       {\relax}        \fi
\providecommand\bibfield[2]{#2}
\providecommand\bibinfo[2]{#2}
\providecommand\natexlab[1]{#1}
\providecommand\showeprint[2][]{arXiv:#2}

\bibitem[\protect\citeauthoryear{API.}{API.}{[n.d.]}]%
        {face++}
\bibfield{author}{\bibinfo{person}{MEGVII Face++ Face~Detection API.}}
  \bibinfo{year}{[n.d.]}\natexlab{}.
\newblock
\newblock
\newblock
\shownote{https://www.faceplusplus.com.}


\bibitem[\protect\citeauthoryear{Ayral, Pedersoli, Bacon, and Granger}{Ayral
  et~al\mbox{.}}{2021}]%
        {ayral2021temporal}
\bibfield{author}{\bibinfo{person}{Th{\'e}o Ayral}, \bibinfo{person}{Marco
  Pedersoli}, \bibinfo{person}{Simon Bacon}, {and} \bibinfo{person}{Eric
  Granger}.} \bibinfo{year}{2021}\natexlab{}.
\newblock \showarticletitle{Temporal stochastic softmax for 3d cnns: An
  application in facial expression recognition}. In
  \bibinfo{booktitle}{\emph{Proceedings of the IEEE/CVF Winter Conference on
  Applications of Computer Vision}}. \bibinfo{pages}{3029--3038}.
\newblock


\bibitem[\protect\citeauthoryear{Beal, Kim, Tzeng, Park, Zhai, and
  Kislyuk}{Beal et~al\mbox{.}}{2020}]%
        {beal2020toward}
\bibfield{author}{\bibinfo{person}{Josh Beal}, \bibinfo{person}{Eric Kim},
  \bibinfo{person}{Eric Tzeng}, \bibinfo{person}{Dong~Huk Park},
  \bibinfo{person}{Andrew Zhai}, {and} \bibinfo{person}{Dmitry Kislyuk}.}
  \bibinfo{year}{2020}\natexlab{}.
\newblock \showarticletitle{Toward Transformer-Based Object Detection}.
\newblock \bibinfo{journal}{\emph{arXiv preprint arXiv:2012.09958}}
  (\bibinfo{year}{2020}).
\newblock


\bibitem[\protect\citeauthoryear{Bertasius, Wang, and Torresani}{Bertasius
  et~al\mbox{.}}{2021}]%
        {bertasius2021space}
\bibfield{author}{\bibinfo{person}{Gedas Bertasius}, \bibinfo{person}{Heng
  Wang}, {and} \bibinfo{person}{Lorenzo Torresani}.}
  \bibinfo{year}{2021}\natexlab{}.
\newblock \showarticletitle{Is space-time attention all you need for video
  understanding}.
\newblock \bibinfo{journal}{\emph{arXiv preprint arXiv:2102.05095}}
  \bibinfo{volume}{2}, \bibinfo{number}{3} (\bibinfo{year}{2021}),
  \bibinfo{pages}{4}.
\newblock


\bibitem[\protect\citeauthoryear{Bisogni, Castiglione, Hossain, Narducci, and
  Umer}{Bisogni et~al\mbox{.}}{2022}]%
        {bisogni2022impact}
\bibfield{author}{\bibinfo{person}{Carmen Bisogni}, \bibinfo{person}{Aniello
  Castiglione}, \bibinfo{person}{Sanoar Hossain}, \bibinfo{person}{Fabio
  Narducci}, {and} \bibinfo{person}{Saiyed Umer}.}
  \bibinfo{year}{2022}\natexlab{}.
\newblock \showarticletitle{Impact of Deep Learning Approaches on Facial
  Expression Recognition in Healthcare Industries}.
\newblock \bibinfo{journal}{\emph{IEEE Transactions on Industrial Informatics}}
  (\bibinfo{year}{2022}).
\newblock
\newblock
\shownote{doi: {10.1109/TII.2022.3141400}.}


\bibitem[\protect\citeauthoryear{Cai, Zheng, Zhang, Li, Cui, and Ye}{Cai
  et~al\mbox{.}}{2016}]%
        {cai2016video}
\bibfield{author}{\bibinfo{person}{Youyi Cai}, \bibinfo{person}{Wenming Zheng},
  \bibinfo{person}{Tong Zhang}, \bibinfo{person}{Qiang Li},
  \bibinfo{person}{Zhen Cui}, {and} \bibinfo{person}{Jiayin Ye}.}
  \bibinfo{year}{2016}\natexlab{}.
\newblock \showarticletitle{Video based emotion recognition using CNN and
  BRNN}. In \bibinfo{booktitle}{\emph{Chinese Conference on Pattern
  Recognition}}. Springer, \bibinfo{pages}{679--691}.
\newblock


\bibitem[\protect\citeauthoryear{Carreira and Zisserman}{Carreira and
  Zisserman}{2017}]%
        {carreira2017quo}
\bibfield{author}{\bibinfo{person}{Joao Carreira} {and} \bibinfo{person}{Andrew
  Zisserman}.} \bibinfo{year}{2017}\natexlab{}.
\newblock \showarticletitle{Quo vadis, action recognition? a new model and the
  kinetics dataset}. In \bibinfo{booktitle}{\emph{proceedings of the IEEE
  Conference on Computer Vision and Pattern Recognition}}.
  \bibinfo{pages}{6299--6308}.
\newblock


\bibitem[\protect\citeauthoryear{Chen, Zhang, Li, and Lee}{Chen
  et~al\mbox{.}}{2020}]%
        {chen2020stcam}
\bibfield{author}{\bibinfo{person}{Weicong Chen}, \bibinfo{person}{Dong Zhang},
  \bibinfo{person}{Ming Li}, {and} \bibinfo{person}{Dah-Jye Lee}.}
  \bibinfo{year}{2020}\natexlab{}.
\newblock \showarticletitle{Stcam: Spatial-temporal and channel attention
  module for dynamic facial expression recognition}.
\newblock \bibinfo{journal}{\emph{IEEE Transactions on Affective Computing}}
  (\bibinfo{year}{2020}).
\newblock
\newblock
\shownote{10.1109/TAFFC.2020.3027340.}


\bibitem[\protect\citeauthoryear{Chung, Gulcehre, Cho, and Bengio}{Chung
  et~al\mbox{.}}{2014}]%
        {chung2014empirical}
\bibfield{author}{\bibinfo{person}{Junyoung Chung}, \bibinfo{person}{Caglar
  Gulcehre}, \bibinfo{person}{KyungHyun Cho}, {and} \bibinfo{person}{Yoshua
  Bengio}.} \bibinfo{year}{2014}\natexlab{}.
\newblock \showarticletitle{Empirical evaluation of gated recurrent neural
  networks on sequence modeling}.
\newblock \bibinfo{journal}{\emph{arXiv preprint arXiv:1412.3555}}
  (\bibinfo{year}{2014}).
\newblock


\bibitem[\protect\citeauthoryear{Darwin}{Darwin}{2015}]%
        {darwin2015expression}
\bibfield{author}{\bibinfo{person}{Charles Darwin}.}
  \bibinfo{year}{2015}\natexlab{}.
\newblock \bibinfo{booktitle}{\emph{The expression of the emotions in man and
  animals}}.
\newblock \bibinfo{publisher}{University of Chicago Press}.
\newblock


\bibitem[\protect\citeauthoryear{Dhall}{Dhall}{2019}]%
        {dhall2019emotiw}
\bibfield{author}{\bibinfo{person}{Abhinav Dhall}.}
  \bibinfo{year}{2019}\natexlab{}.
\newblock \showarticletitle{Emotiw 2019: Automatic emotion, engagement and
  cohesion prediction tasks}. In \bibinfo{booktitle}{\emph{International
  Conference on Multimodal Interaction}}. \bibinfo{pages}{546--550}.
\newblock


\bibitem[\protect\citeauthoryear{Dosovitskiy, Beyer, Kolesnikov, Weissenborn,
  Zhai, Unterthiner, Dehghani, Minderer, Heigold, Gelly,
  et~al\mbox{.}}{Dosovitskiy et~al\mbox{.}}{2020}]%
        {dosovitskiy2020image}
\bibfield{author}{\bibinfo{person}{Alexey Dosovitskiy}, \bibinfo{person}{Lucas
  Beyer}, \bibinfo{person}{Alexander Kolesnikov}, \bibinfo{person}{Dirk
  Weissenborn}, \bibinfo{person}{Xiaohua Zhai}, \bibinfo{person}{Thomas
  Unterthiner}, \bibinfo{person}{Mostafa Dehghani}, \bibinfo{person}{Matthias
  Minderer}, \bibinfo{person}{Georg Heigold}, \bibinfo{person}{Sylvain Gelly},
  {et~al\mbox{.}}} \bibinfo{year}{2020}\natexlab{}.
\newblock \showarticletitle{An image is worth 16x16 words: Transformers for
  image recognition at scale}.
\newblock \bibinfo{journal}{\emph{arXiv preprint arXiv:2010.11929}}
  (\bibinfo{year}{2020}).
\newblock


\bibitem[\protect\citeauthoryear{Esser, Rombach, and Ommer}{Esser
  et~al\mbox{.}}{2021}]%
        {esser2021taming}
\bibfield{author}{\bibinfo{person}{Patrick Esser}, \bibinfo{person}{Robin
  Rombach}, {and} \bibinfo{person}{Bjorn Ommer}.}
  \bibinfo{year}{2021}\natexlab{}.
\newblock \showarticletitle{Taming transformers for high-resolution image
  synthesis}. In \bibinfo{booktitle}{\emph{Proceedings of the IEEE/CVF
  Conference on Computer Vision and Pattern Recognition}}.
  \bibinfo{pages}{12873--12883}.
\newblock


\bibitem[\protect\citeauthoryear{Guo, Zhang, Hu, He, and Gao}{Guo
  et~al\mbox{.}}{2016}]%
        {guo2016ms}
\bibfield{author}{\bibinfo{person}{Yandong Guo}, \bibinfo{person}{Lei Zhang},
  \bibinfo{person}{Yuxiao Hu}, \bibinfo{person}{Xiaodong He}, {and}
  \bibinfo{person}{Jianfeng Gao}.} \bibinfo{year}{2016}\natexlab{}.
\newblock \showarticletitle{Ms-celeb-1m: A dataset and benchmark for
  large-scale face recognition}. In \bibinfo{booktitle}{\emph{European
  Conference on Computer Vision}}. Springer, \bibinfo{pages}{87--102}.
\newblock


\bibitem[\protect\citeauthoryear{Hara, Kataoka, and Satoh}{Hara
  et~al\mbox{.}}{2018}]%
        {hara2018can}
\bibfield{author}{\bibinfo{person}{Kensho Hara}, \bibinfo{person}{Hirokatsu
  Kataoka}, {and} \bibinfo{person}{Yutaka Satoh}.}
  \bibinfo{year}{2018}\natexlab{}.
\newblock \showarticletitle{Can spatiotemporal 3d cnns retrace the history of
  2d cnns and imagenet?}. In \bibinfo{booktitle}{\emph{Proceedings of the IEEE
  conference on Computer Vision and Pattern Recognition}}.
  \bibinfo{pages}{6546--6555}.
\newblock


\bibitem[\protect\citeauthoryear{He, Zhang, Ren, and Sun}{He
  et~al\mbox{.}}{2016}]%
        {he2016deep}
\bibfield{author}{\bibinfo{person}{Kaiming He}, \bibinfo{person}{Xiangyu
  Zhang}, \bibinfo{person}{Shaoqing Ren}, {and} \bibinfo{person}{Jian Sun}.}
  \bibinfo{year}{2016}\natexlab{}.
\newblock \showarticletitle{Deep residual learning for image recognition}. In
  \bibinfo{booktitle}{\emph{Proceedings of the IEEE Conference on Computer
  Vision and Pattern Recognition}}. \bibinfo{pages}{770--778}.
\newblock


\bibitem[\protect\citeauthoryear{Hendrycks and Gimpel}{Hendrycks and
  Gimpel}{2016}]%
        {hendrycks2016gaussian}
\bibfield{author}{\bibinfo{person}{Dan Hendrycks} {and} \bibinfo{person}{Kevin
  Gimpel}.} \bibinfo{year}{2016}\natexlab{}.
\newblock \showarticletitle{Gaussian error linear units (gelus)}.
\newblock \bibinfo{journal}{\emph{arXiv preprint arXiv:1606.08415}}
  (\bibinfo{year}{2016}).
\newblock


\bibitem[\protect\citeauthoryear{Hochreiter and Schmidhuber}{Hochreiter and
  Schmidhuber}{1997}]%
        {hochreiter1997long}
\bibfield{author}{\bibinfo{person}{Sepp Hochreiter} {and}
  \bibinfo{person}{J{\"u}rgen Schmidhuber}.} \bibinfo{year}{1997}\natexlab{}.
\newblock \showarticletitle{Long short-term memory}.
\newblock \bibinfo{journal}{\emph{Neural Computation}} \bibinfo{volume}{9},
  \bibinfo{number}{8} (\bibinfo{year}{1997}), \bibinfo{pages}{1735--1780}.
\newblock


\bibitem[\protect\citeauthoryear{Huang, Wang, and Ying}{Huang
  et~al\mbox{.}}{2010}]%
        {huang2010new}
\bibfield{author}{\bibinfo{person}{Ming-Wei Huang}, \bibinfo{person}{Zhe-wei
  Wang}, {and} \bibinfo{person}{Zi-Lu Ying}.} \bibinfo{year}{2010}\natexlab{}.
\newblock \showarticletitle{A new method for facial expression recognition
  based on sparse representation plus LBP}. In
  \bibinfo{booktitle}{\emph{International Congress on Image and Signal
  Processing}}, Vol.~\bibinfo{volume}{4}. IEEE, \bibinfo{pages}{1750--1754}.
\newblock


\bibitem[\protect\citeauthoryear{Huang, Huang, Wang, and Jiang}{Huang
  et~al\mbox{.}}{2021}]%
        {huang2021facial}
\bibfield{author}{\bibinfo{person}{Qionghao Huang}, \bibinfo{person}{Changqin
  Huang}, \bibinfo{person}{Xizhe Wang}, {and} \bibinfo{person}{Fan Jiang}.}
  \bibinfo{year}{2021}\natexlab{}.
\newblock \showarticletitle{Facial expression recognition with grid-wise
  attention and visual transformer}.
\newblock \bibinfo{journal}{\emph{Information Sciences}}  \bibinfo{volume}{580}
  (\bibinfo{year}{2021}), \bibinfo{pages}{35--54}.
\newblock


\bibitem[\protect\citeauthoryear{Jiang, Zong, Zheng, Tang, Xia, Lu, and
  Liu}{Jiang et~al\mbox{.}}{2020}]%
        {jiang2020dfew}
\bibfield{author}{\bibinfo{person}{Xingxun Jiang}, \bibinfo{person}{Yuan Zong},
  \bibinfo{person}{Wenming Zheng}, \bibinfo{person}{Chuangao Tang},
  \bibinfo{person}{Wanchuang Xia}, \bibinfo{person}{Cheng Lu}, {and}
  \bibinfo{person}{Jiateng Liu}.} \bibinfo{year}{2020}\natexlab{}.
\newblock \showarticletitle{Dfew: A large-scale database for recognizing
  dynamic facial expressions in the wild}. In
  \bibinfo{booktitle}{\emph{Proceedings of the 28th ACM International
  Conference on Multimedia}}. \bibinfo{pages}{2881--2889}.
\newblock


\bibitem[\protect\citeauthoryear{Jung, Lee, Yim, Park, and Kim}{Jung
  et~al\mbox{.}}{2015}]%
        {jung2015joint}
\bibfield{author}{\bibinfo{person}{Heechul Jung}, \bibinfo{person}{Sihaeng
  Lee}, \bibinfo{person}{Junho Yim}, \bibinfo{person}{Sunjeong Park}, {and}
  \bibinfo{person}{Junmo Kim}.} \bibinfo{year}{2015}\natexlab{}.
\newblock \showarticletitle{Joint fine-tuning in deep neural networks for
  facial expression recognition}. In \bibinfo{booktitle}{\emph{Proceedings of
  the IEEE International Conference on Computer Vision}}.
  \bibinfo{pages}{2983--2991}.
\newblock


\bibitem[\protect\citeauthoryear{Kim, Baddar, Jang, and Ro}{Kim
  et~al\mbox{.}}{2017}]%
        {kim2017multi}
\bibfield{author}{\bibinfo{person}{Dae~Hoe Kim}, \bibinfo{person}{Wissam~J
  Baddar}, \bibinfo{person}{Jinhyeok Jang}, {and} \bibinfo{person}{Yong~Man
  Ro}.} \bibinfo{year}{2017}\natexlab{}.
\newblock \showarticletitle{Multi-objective based spatio-temporal feature
  representation learning robust to expression intensity variations for facial
  expression recognition}.
\newblock \bibinfo{journal}{\emph{IEEE Transactions on Affective Computing}}
  \bibinfo{volume}{10}, \bibinfo{number}{2} (\bibinfo{year}{2017}),
  \bibinfo{pages}{223--236}.
\newblock


\bibitem[\protect\citeauthoryear{King}{King}{2009}]%
        {dlib09}
\bibfield{author}{\bibinfo{person}{Davis~E. King}.}
  \bibinfo{year}{2009}\natexlab{}.
\newblock \showarticletitle{Dlib-ml: A Machine Learning Toolkit}.
\newblock \bibinfo{journal}{\emph{Journal of Machine Learning Research}}
  \bibinfo{volume}{10} (\bibinfo{year}{2009}), \bibinfo{pages}{1755--1758}.
\newblock


\bibitem[\protect\citeauthoryear{Kobayashi}{Kobayashi}{2019}]%
        {kobayashi2019large}
\bibfield{author}{\bibinfo{person}{Takumi Kobayashi}.}
  \bibinfo{year}{2019}\natexlab{}.
\newblock \showarticletitle{Large Margin In Softmax Cross-Entropy Loss.}. In
  \bibinfo{booktitle}{\emph{British Machine Vision Conference}}.
  \bibinfo{pages}{139}.
\newblock


\bibitem[\protect\citeauthoryear{Kollias, Tzirakis, Nicolaou, Papaioannou,
  Zhao, Schuller, Kotsia, and Zafeiriou}{Kollias et~al\mbox{.}}{2019}]%
        {kollias2019deep}
\bibfield{author}{\bibinfo{person}{Dimitrios Kollias},
  \bibinfo{person}{Panagiotis Tzirakis}, \bibinfo{person}{Mihalis~A Nicolaou},
  \bibinfo{person}{Athanasios Papaioannou}, \bibinfo{person}{Guoying Zhao},
  \bibinfo{person}{Bj{\"o}rn Schuller}, \bibinfo{person}{Irene Kotsia}, {and}
  \bibinfo{person}{Stefanos Zafeiriou}.} \bibinfo{year}{2019}\natexlab{}.
\newblock \showarticletitle{Deep affect prediction in-the-wild: Aff-wild
  database and challenge, deep architectures, and beyond}.
\newblock \bibinfo{journal}{\emph{International Journal of Computer Vision}}
  \bibinfo{volume}{127}, \bibinfo{number}{6} (\bibinfo{year}{2019}),
  \bibinfo{pages}{907--929}.
\newblock


\bibitem[\protect\citeauthoryear{Kuo, Lai, and Sarkis}{Kuo
  et~al\mbox{.}}{2018}]%
        {kuo2018compact}
\bibfield{author}{\bibinfo{person}{Chieh-Ming Kuo}, \bibinfo{person}{Shang-Hong
  Lai}, {and} \bibinfo{person}{Michel Sarkis}.}
  \bibinfo{year}{2018}\natexlab{}.
\newblock \showarticletitle{A compact deep learning model for robust facial
  expression recognition}. In \bibinfo{booktitle}{\emph{Proceedings of the IEEE
  Conference on Computer Vision and Pattern Recognition Workshops}}.
  \bibinfo{pages}{2121--2129}.
\newblock


\bibitem[\protect\citeauthoryear{Lee, Kim, Kim, Park, and Sohn}{Lee
  et~al\mbox{.}}{2019}]%
        {lee2019context}
\bibfield{author}{\bibinfo{person}{Jiyoung Lee}, \bibinfo{person}{Seungryong
  Kim}, \bibinfo{person}{Sunok Kim}, \bibinfo{person}{Jungin Park}, {and}
  \bibinfo{person}{Kwanghoon Sohn}.} \bibinfo{year}{2019}\natexlab{}.
\newblock \showarticletitle{Context-aware emotion recognition networks}. In
  \bibinfo{booktitle}{\emph{Proceedings of the IEEE/CVF International
  Conference on Computer Vision}}. \bibinfo{pages}{10143--10152}.
\newblock


\bibitem[\protect\citeauthoryear{Lee, Baddar, and Ro}{Lee
  et~al\mbox{.}}{2016}]%
        {lee2016collaborative}
\bibfield{author}{\bibinfo{person}{Seung~Ho Lee}, \bibinfo{person}{Wissam~J
  Baddar}, {and} \bibinfo{person}{Yong~Man Ro}.}
  \bibinfo{year}{2016}\natexlab{}.
\newblock \showarticletitle{Collaborative expression representation using peak
  expression and intra class variation face images for practical
  subject-independent emotion recognition in videos}.
\newblock \bibinfo{journal}{\emph{Pattern Recognition}}  \bibinfo{volume}{54}
  (\bibinfo{year}{2016}), \bibinfo{pages}{52--67}.
\newblock


\bibitem[\protect\citeauthoryear{Li, Sui, Zhao, Zha, and Wu}{Li
  et~al\mbox{.}}{2021}]%
        {li2021mvt}
\bibfield{author}{\bibinfo{person}{Hanting Li}, \bibinfo{person}{Mingzhe Sui},
  \bibinfo{person}{Feng Zhao}, \bibinfo{person}{Zhengjun Zha}, {and}
  \bibinfo{person}{Feng Wu}.} \bibinfo{year}{2021}\natexlab{}.
\newblock \showarticletitle{Mvt: Mask vision transformer for facial expression
  recognition in the wild}.
\newblock \bibinfo{journal}{\emph{arXiv preprint arXiv:2106.04520}}
  (\bibinfo{year}{2021}).
\newblock


\bibitem[\protect\citeauthoryear{Li and Deng}{Li and Deng}{2018}]%
        {li2017reliable}
\bibfield{author}{\bibinfo{person}{Shan Li} {and} \bibinfo{person}{Weihong
  Deng}.} \bibinfo{year}{2018}\natexlab{}.
\newblock \showarticletitle{Reliable crowdsourcing and deep locality-preserving
  learning for unconstrained facial expression recognition}.
\newblock \bibinfo{journal}{\emph{IEEE Transactions on Image Processing}}
  \bibinfo{volume}{28}, \bibinfo{number}{1} (\bibinfo{year}{2018}),
  \bibinfo{pages}{356--370}.
\newblock


\bibitem[\protect\citeauthoryear{Li and Deng}{Li and Deng}{2020}]%
        {li2020deep}
\bibfield{author}{\bibinfo{person}{Shan Li} {and} \bibinfo{person}{Weihong
  Deng}.} \bibinfo{year}{2020}\natexlab{}.
\newblock \showarticletitle{Deep facial expression recognition: A survey}.
\newblock \bibinfo{journal}{\emph{IEEE Transactions on Affective Computing}}
  (\bibinfo{year}{2020}).
\newblock
\newblock
\shownote{doi: {10.1109/TAFFC.2020.2981446.}}


\bibitem[\protect\citeauthoryear{Liang, Liang, Yu, and Zhang}{Liang
  et~al\mbox{.}}{2020}]%
        {liang2020deep}
\bibfield{author}{\bibinfo{person}{Dandan Liang}, \bibinfo{person}{Huagang
  Liang}, \bibinfo{person}{Zhenbo Yu}, {and} \bibinfo{person}{Yipu Zhang}.}
  \bibinfo{year}{2020}\natexlab{}.
\newblock \showarticletitle{Deep convolutional BiLSTM fusion network for facial
  expression recognition}.
\newblock \bibinfo{journal}{\emph{The Visual Computer}} \bibinfo{volume}{36},
  \bibinfo{number}{3} (\bibinfo{year}{2020}), \bibinfo{pages}{499--508}.
\newblock


\bibitem[\protect\citeauthoryear{Liu, Li, Shan, Wang, and Chen}{Liu
  et~al\mbox{.}}{2014}]%
        {liu2014deeply}
\bibfield{author}{\bibinfo{person}{Mengyi Liu}, \bibinfo{person}{Shaoxin Li},
  \bibinfo{person}{Shiguang Shan}, \bibinfo{person}{Ruiping Wang}, {and}
  \bibinfo{person}{Xilin Chen}.} \bibinfo{year}{2014}\natexlab{}.
\newblock \showarticletitle{Deeply learning deformable facial action parts
  model for dynamic expression analysis}. In \bibinfo{booktitle}{\emph{Asian
  Conference on Computer Vision}}. Springer, \bibinfo{pages}{143--157}.
\newblock


\bibitem[\protect\citeauthoryear{Liu, Wen, Yu, and Yang}{Liu
  et~al\mbox{.}}{2016}]%
        {liu2016large}
\bibfield{author}{\bibinfo{person}{Weiyang Liu}, \bibinfo{person}{Yandong Wen},
  \bibinfo{person}{Zhiding Yu}, {and} \bibinfo{person}{Meng Yang}.}
  \bibinfo{year}{2016}\natexlab{}.
\newblock \showarticletitle{Large-margin softmax loss for convolutional neural
  networks.}. In \bibinfo{booktitle}{\emph{International Conference on Machine
  Learning}}, Vol.~\bibinfo{volume}{2}. \bibinfo{pages}{7}.
\newblock


\bibitem[\protect\citeauthoryear{Liu, Kan, Wu, Shan, and Chen}{Liu
  et~al\mbox{.}}{2017a}]%
        {liu2017viplfacenet}
\bibfield{author}{\bibinfo{person}{Xin Liu}, \bibinfo{person}{Meina Kan},
  \bibinfo{person}{Wanglong Wu}, \bibinfo{person}{Shiguang Shan}, {and}
  \bibinfo{person}{Xilin Chen}.} \bibinfo{year}{2017}\natexlab{a}.
\newblock \showarticletitle{VIPLFaceNet: an open source deep face recognition
  SDK}.
\newblock \bibinfo{journal}{\emph{Frontiers of Computer Science}}
  \bibinfo{volume}{11}, \bibinfo{number}{2} (\bibinfo{year}{2017}),
  \bibinfo{pages}{208--218}.
\newblock


\bibitem[\protect\citeauthoryear{Liu, Feng, Yuan, Zhou, Wang, Qin, and Luo}{Liu
  et~al\mbox{.}}{2022}]%
        {liu2022clip}
\bibfield{author}{\bibinfo{person}{Yuanyuan Liu}, \bibinfo{person}{Chuanxu
  Feng}, \bibinfo{person}{Xiaohui Yuan}, \bibinfo{person}{Lin Zhou},
  \bibinfo{person}{Wenbin Wang}, \bibinfo{person}{Jie Qin}, {and}
  \bibinfo{person}{Zhongwen Luo}.} \bibinfo{year}{2022}\natexlab{}.
\newblock \showarticletitle{Clip-aware Expressive Feature Learning for
  Video-based Facial Expression Recognition}.
\newblock \bibinfo{journal}{\emph{Information Sciences}}
  (\bibinfo{year}{2022}).
\newblock
\newblock
\shownote{doi: {10.1016/j.ins.2022.03.062}.}


\bibitem[\protect\citeauthoryear{Liu, Yuan, Gong, Xie, Fang, and Luo}{Liu
  et~al\mbox{.}}{2018}]%
        {liu2018conditional}
\bibfield{author}{\bibinfo{person}{Yuanyuan Liu}, \bibinfo{person}{Xiaohui
  Yuan}, \bibinfo{person}{Xi Gong}, \bibinfo{person}{Zhong Xie},
  \bibinfo{person}{Fang Fang}, {and} \bibinfo{person}{Zhongwen Luo}.}
  \bibinfo{year}{2018}\natexlab{}.
\newblock \showarticletitle{Conditional convolution neural network enhanced
  random forest for facial expression recognition}.
\newblock \bibinfo{journal}{\emph{Pattern Recognition}}  \bibinfo{volume}{84}
  (\bibinfo{year}{2018}), \bibinfo{pages}{251--261}.
\newblock


\bibitem[\protect\citeauthoryear{Liu, Wu, Cao, Chen, Xu, Zhang, Zhou, and
  Mao}{Liu et~al\mbox{.}}{2017b}]%
        {liu2017facial}
\bibfield{author}{\bibinfo{person}{Zhentao Liu}, \bibinfo{person}{Min Wu},
  \bibinfo{person}{Weihua Cao}, \bibinfo{person}{Luefeng Chen},
  \bibinfo{person}{Jianping Xu}, \bibinfo{person}{Ri Zhang},
  \bibinfo{person}{Mengtian Zhou}, {and} \bibinfo{person}{Junwei Mao}.}
  \bibinfo{year}{2017}\natexlab{b}.
\newblock \showarticletitle{A facial expression emotion recognition based
  human-robot interaction system}.
\newblock \bibinfo{journal}{\emph{IEEE/CAA Journal of Automatica Sinica}}
  \bibinfo{volume}{4}, \bibinfo{number}{4} (\bibinfo{year}{2017}),
  \bibinfo{pages}{668--676}.
\newblock


\bibitem[\protect\citeauthoryear{Lucey, Cohn, Kanade, Saragih, Ambadar, and
  Matthews}{Lucey et~al\mbox{.}}{2010}]%
        {lucey2010extended}
\bibfield{author}{\bibinfo{person}{Patrick Lucey}, \bibinfo{person}{Jeffrey~F
  Cohn}, \bibinfo{person}{Takeo Kanade}, \bibinfo{person}{Jason Saragih},
  \bibinfo{person}{Zara Ambadar}, {and} \bibinfo{person}{Iain Matthews}.}
  \bibinfo{year}{2010}\natexlab{}.
\newblock \showarticletitle{The extended cohn-kanade dataset (ck+): A complete
  dataset for action unit and emotion-specified expression}. In
  \bibinfo{booktitle}{\emph{IEEE Computer Society Conference on Computer Vision
  and Pattern Recognition-Workshops}}. IEEE, \bibinfo{pages}{94--101}.
\newblock


\bibitem[\protect\citeauthoryear{Ma, Sun, and Li}{Ma et~al\mbox{.}}{2021}]%
        {ma2021facial}
\bibfield{author}{\bibinfo{person}{Fuyan Ma}, \bibinfo{person}{Bin Sun}, {and}
  \bibinfo{person}{Shutao Li}.} \bibinfo{year}{2021}\natexlab{}.
\newblock \showarticletitle{Facial Expression Recognition with Visual
  Transformers and Attentional Selective Fusion}.
\newblock \bibinfo{journal}{\emph{IEEE Transactions on Affective Computing}}
  (\bibinfo{year}{2021}).
\newblock
\newblock
\shownote{doi: {10.1109/TAFFC.2021.3122146}.}


\bibitem[\protect\citeauthoryear{Meng, Peng, Wang, and Qiao}{Meng
  et~al\mbox{.}}{2019}]%
        {meng2019frame}
\bibfield{author}{\bibinfo{person}{Debin Meng}, \bibinfo{person}{Xiaojiang
  Peng}, \bibinfo{person}{Kai Wang}, {and} \bibinfo{person}{Yu Qiao}.}
  \bibinfo{year}{2019}\natexlab{}.
\newblock \showarticletitle{Frame attention networks for facial expression
  recognition in videos}. In \bibinfo{booktitle}{\emph{IEEE International
  Conference on Image Processing}}. IEEE, \bibinfo{pages}{3866--3870}.
\newblock


\bibitem[\protect\citeauthoryear{Paszke, Gross, Massa, Lerer, Bradbury, Chanan,
  Killeen, Lin, Gimelshein, Antiga, et~al\mbox{.}}{Paszke
  et~al\mbox{.}}{2019}]%
        {paszke2019pytorch}
\bibfield{author}{\bibinfo{person}{Adam Paszke}, \bibinfo{person}{Sam Gross},
  \bibinfo{person}{Francisco Massa}, \bibinfo{person}{Adam Lerer},
  \bibinfo{person}{James Bradbury}, \bibinfo{person}{Gregory Chanan},
  \bibinfo{person}{Trevor Killeen}, \bibinfo{person}{Zeming Lin},
  \bibinfo{person}{Natalia Gimelshein}, \bibinfo{person}{Luca Antiga},
  {et~al\mbox{.}}} \bibinfo{year}{2019}\natexlab{}.
\newblock \showarticletitle{Pytorch: An imperative style, high-performance deep
  learning library}.
\newblock \bibinfo{journal}{\emph{arXiv preprint arXiv:1912.01703}}
  (\bibinfo{year}{2019}).
\newblock


\bibitem[\protect\citeauthoryear{Qiu, Yao, and Mei}{Qiu et~al\mbox{.}}{2017}]%
        {qiu2017learning}
\bibfield{author}{\bibinfo{person}{Zhaofan Qiu}, \bibinfo{person}{Ting Yao},
  {and} \bibinfo{person}{Tao Mei}.} \bibinfo{year}{2017}\natexlab{}.
\newblock \showarticletitle{Learning spatio-temporal representation with
  pseudo-3d residual networks}. In \bibinfo{booktitle}{\emph{proceedings of the
  IEEE International Conference on Computer Vision}}.
  \bibinfo{pages}{5533--5541}.
\newblock


\bibitem[\protect\citeauthoryear{Ruder}{Ruder}{2016}]%
        {ruder2016overview}
\bibfield{author}{\bibinfo{person}{Sebastian Ruder}.}
  \bibinfo{year}{2016}\natexlab{}.
\newblock \showarticletitle{An overview of gradient descent optimization
  algorithms}.
\newblock \bibinfo{journal}{\emph{arXiv preprint arXiv:1609.04747}}
  (\bibinfo{year}{2016}).
\newblock


\bibitem[\protect\citeauthoryear{Selvaraju, Cogswell, Das, Vedantam, Parikh,
  and Batra}{Selvaraju et~al\mbox{.}}{2017}]%
        {selvaraju2017grad}
\bibfield{author}{\bibinfo{person}{Ramprasaath~R Selvaraju},
  \bibinfo{person}{Michael Cogswell}, \bibinfo{person}{Abhishek Das},
  \bibinfo{person}{Ramakrishna Vedantam}, \bibinfo{person}{Devi Parikh}, {and}
  \bibinfo{person}{Dhruv Batra}.} \bibinfo{year}{2017}\natexlab{}.
\newblock \showarticletitle{Grad-cam: Visual explanations from deep networks
  via gradient-based localization}. In \bibinfo{booktitle}{\emph{Proceedings of
  the IEEE International Conference on Computer Vision}}.
  \bibinfo{pages}{618--626}.
\newblock


\bibitem[\protect\citeauthoryear{Sharir, Noy, and Zelnik-Manor}{Sharir
  et~al\mbox{.}}{2021}]%
        {sharir2021image}
\bibfield{author}{\bibinfo{person}{Gilad Sharir}, \bibinfo{person}{Asaf Noy},
  {and} \bibinfo{person}{Lihi Zelnik-Manor}.} \bibinfo{year}{2021}\natexlab{}.
\newblock \showarticletitle{An image is worth 16x16 words, what is a video
  worth?}
\newblock \bibinfo{journal}{\emph{arXiv preprint arXiv:2103.13915}}
  (\bibinfo{year}{2021}).
\newblock


\bibitem[\protect\citeauthoryear{Simonyan and Zisserman}{Simonyan and
  Zisserman}{2014}]%
        {simonyan2014very}
\bibfield{author}{\bibinfo{person}{Karen Simonyan} {and}
  \bibinfo{person}{Andrew Zisserman}.} \bibinfo{year}{2014}\natexlab{}.
\newblock \showarticletitle{Very deep convolutional networks for large-scale
  image recognition}.
\newblock \bibinfo{journal}{\emph{arXiv preprint arXiv:1409.1556}}
  (\bibinfo{year}{2014}).
\newblock


\bibitem[\protect\citeauthoryear{Tomar}{Tomar}{2006}]%
        {tomar2006converting}
\bibfield{author}{\bibinfo{person}{Suramya Tomar}.}
  \bibinfo{year}{2006}\natexlab{}.
\newblock \showarticletitle{Converting video formats with FFmpeg}.
\newblock \bibinfo{journal}{\emph{Linux Journal}} \bibinfo{volume}{2006},
  \bibinfo{number}{146} (\bibinfo{year}{2006}), \bibinfo{pages}{10}.
\newblock


\bibitem[\protect\citeauthoryear{Tran, Bourdev, Fergus, Torresani, and
  Paluri}{Tran et~al\mbox{.}}{2015}]%
        {tran2015learning}
\bibfield{author}{\bibinfo{person}{Du Tran}, \bibinfo{person}{Lubomir Bourdev},
  \bibinfo{person}{Rob Fergus}, \bibinfo{person}{Lorenzo Torresani}, {and}
  \bibinfo{person}{Manohar Paluri}.} \bibinfo{year}{2015}\natexlab{}.
\newblock \showarticletitle{Learning spatiotemporal features with 3d
  convolutional networks}. In \bibinfo{booktitle}{\emph{Proceedings of the IEEE
  international Conference on Computer Vision}}. \bibinfo{pages}{4489--4497}.
\newblock


\bibitem[\protect\citeauthoryear{Tran, Wang, Torresani, Ray, LeCun, and
  Paluri}{Tran et~al\mbox{.}}{2018}]%
        {tran2018closer}
\bibfield{author}{\bibinfo{person}{Du Tran}, \bibinfo{person}{Heng Wang},
  \bibinfo{person}{Lorenzo Torresani}, \bibinfo{person}{Jamie Ray},
  \bibinfo{person}{Yann LeCun}, {and} \bibinfo{person}{Manohar Paluri}.}
  \bibinfo{year}{2018}\natexlab{}.
\newblock \showarticletitle{A closer look at spatiotemporal convolutions for
  action recognition}. In \bibinfo{booktitle}{\emph{Proceedings of the IEEE
  conference on Computer Vision and Pattern Recognition}}.
  \bibinfo{pages}{6450--6459}.
\newblock


\bibitem[\protect\citeauthoryear{Valstar, Pantic, et~al\mbox{.}}{Valstar
  et~al\mbox{.}}{2010}]%
        {valstar2010induced}
\bibfield{author}{\bibinfo{person}{Michel Valstar}, \bibinfo{person}{Maja
  Pantic}, {et~al\mbox{.}}} \bibinfo{year}{2010}\natexlab{}.
\newblock \showarticletitle{Induced disgust, happiness and surprise: an
  addition to the mmi facial expression database}. In
  \bibinfo{booktitle}{\emph{Proceedings of International Workshop on EMOTION:
  Corpora for Research on Emotion and Affect}}. Paris, France.,
  \bibinfo{pages}{65}.
\newblock


\bibitem[\protect\citeauthoryear{Van~der Maaten and Hinton}{Van~der Maaten and
  Hinton}{2008}]%
        {van2008visualizing}
\bibfield{author}{\bibinfo{person}{Laurens Van~der Maaten} {and}
  \bibinfo{person}{Geoffrey Hinton}.} \bibinfo{year}{2008}\natexlab{}.
\newblock \showarticletitle{Visualizing data using t-SNE.}
\newblock \bibinfo{journal}{\emph{Journal of Machine Learning Research}}
  \bibinfo{volume}{9}, \bibinfo{number}{11} (\bibinfo{year}{2008}).
\newblock


\bibitem[\protect\citeauthoryear{Vielzeuf, Pateux, and Jurie}{Vielzeuf
  et~al\mbox{.}}{2017}]%
        {vielzeuf2017temporal}
\bibfield{author}{\bibinfo{person}{Valentin Vielzeuf},
  \bibinfo{person}{St{\'e}phane Pateux}, {and}
  \bibinfo{person}{Fr{\'e}d{\'e}ric Jurie}.} \bibinfo{year}{2017}\natexlab{}.
\newblock \showarticletitle{Temporal multimodal fusion for video emotion
  classification in the wild}. In \bibinfo{booktitle}{\emph{Proceedings of the
  19th ACM International Conference on Multimodal Interaction}}.
  \bibinfo{pages}{569--576}.
\newblock


\bibitem[\protect\citeauthoryear{Wen, Zhang, Li, and Qiao}{Wen
  et~al\mbox{.}}{2016}]%
        {wen2016discriminative}
\bibfield{author}{\bibinfo{person}{Yandong Wen}, \bibinfo{person}{Kaipeng
  Zhang}, \bibinfo{person}{Zhifeng Li}, {and} \bibinfo{person}{Yu Qiao}.}
  \bibinfo{year}{2016}\natexlab{}.
\newblock \showarticletitle{A discriminative feature learning approach for deep
  face recognition}. In \bibinfo{booktitle}{\emph{European Conference on
  Computer Vision}}. Springer, \bibinfo{pages}{499--515}.
\newblock


\bibitem[\protect\citeauthoryear{Wilhelm}{Wilhelm}{2019}]%
        {wilhelm2019towards}
\bibfield{author}{\bibinfo{person}{Torsten Wilhelm}.}
  \bibinfo{year}{2019}\natexlab{}.
\newblock \showarticletitle{Towards facial expression analysis in a driver
  assistance system}. In \bibinfo{booktitle}{\emph{2019 14th IEEE International
  Conference on Automatic Face \& Gesture Recognition (FG 2019)}}. IEEE,
  \bibinfo{pages}{1--4}.
\newblock


\bibitem[\protect\citeauthoryear{Yang, Ciftci, and Yin}{Yang
  et~al\mbox{.}}{2018}]%
        {yang2018facial}
\bibfield{author}{\bibinfo{person}{Huiyuan Yang}, \bibinfo{person}{Umur
  Ciftci}, {and} \bibinfo{person}{Lijun Yin}.} \bibinfo{year}{2018}\natexlab{}.
\newblock \showarticletitle{Facial expression recognition by de-expression
  residue learning}. In \bibinfo{booktitle}{\emph{Proceedings of the IEEE
  Conference on Computer Vision and Pattern Recognition}}.
  \bibinfo{pages}{2168--2177}.
\newblock


\bibitem[\protect\citeauthoryear{Yang, Quan, Nie, and Yang}{Yang
  et~al\mbox{.}}{2020}]%
        {yang2020transpose}
\bibfield{author}{\bibinfo{person}{Sen Yang}, \bibinfo{person}{Zhibin Quan},
  \bibinfo{person}{Mu Nie}, {and} \bibinfo{person}{Wankou Yang}.}
  \bibinfo{year}{2020}\natexlab{}.
\newblock \showarticletitle{TransPose: Towards Explainable Human Pose
  Estimation by Transformer}.
\newblock \bibinfo{journal}{\emph{arXiv preprint arXiv:2012.14214}}
  (\bibinfo{year}{2020}).
\newblock


\bibitem[\protect\citeauthoryear{Yu, Zheng, Peng, Dong, and Du}{Yu
  et~al\mbox{.}}{2020}]%
        {yu2020facial}
\bibfield{author}{\bibinfo{person}{Mingjing Yu}, \bibinfo{person}{Huicheng
  Zheng}, \bibinfo{person}{Zhifeng Peng}, \bibinfo{person}{Jiayu Dong}, {and}
  \bibinfo{person}{Heran Du}.} \bibinfo{year}{2020}\natexlab{}.
\newblock \showarticletitle{Facial expression recognition based on a multi-task
  global-local network}.
\newblock \bibinfo{journal}{\emph{Pattern Recognition Letters}}
  \bibinfo{volume}{131} (\bibinfo{year}{2020}), \bibinfo{pages}{166--171}.
\newblock


\bibitem[\protect\citeauthoryear{Zhao, Huang, Taini, Li, and
  Pietik{\"a}Inen}{Zhao et~al\mbox{.}}{2011}]%
        {zhao2011facial}
\bibfield{author}{\bibinfo{person}{Guoying Zhao}, \bibinfo{person}{Xiaohua
  Huang}, \bibinfo{person}{Matti Taini}, \bibinfo{person}{Stan~Z Li}, {and}
  \bibinfo{person}{Matti Pietik{\"a}Inen}.} \bibinfo{year}{2011}\natexlab{}.
\newblock \showarticletitle{Facial expression recognition from near-infrared
  videos}.
\newblock \bibinfo{journal}{\emph{Image and Vision Computing}}
  \bibinfo{volume}{29}, \bibinfo{number}{9} (\bibinfo{year}{2011}),
  \bibinfo{pages}{607--619}.
\newblock


\bibitem[\protect\citeauthoryear{Zhao, Liang, Liu, Li, Han, Vasconcelos, and
  Yan}{Zhao et~al\mbox{.}}{2016}]%
        {zhao2016peak}
\bibfield{author}{\bibinfo{person}{Xiangyun Zhao}, \bibinfo{person}{Xiaodan
  Liang}, \bibinfo{person}{Luoqi Liu}, \bibinfo{person}{Teng Li},
  \bibinfo{person}{Yugang Han}, \bibinfo{person}{Nuno Vasconcelos}, {and}
  \bibinfo{person}{Shuicheng Yan}.} \bibinfo{year}{2016}\natexlab{}.
\newblock \showarticletitle{Peak-piloted deep network for facial expression
  recognition}. In \bibinfo{booktitle}{\emph{European Conference on Computer
  Vision}}. Springer, \bibinfo{pages}{425--442}.
\newblock


\bibitem[\protect\citeauthoryear{Zhao and Liu}{Zhao and Liu}{2021}]%
        {zhao2021former}
\bibfield{author}{\bibinfo{person}{Zengqun Zhao} {and}
  \bibinfo{person}{Qingshan Liu}.} \bibinfo{year}{2021}\natexlab{}.
\newblock \showarticletitle{Former-DFER: Dynamic Facial Expression Recognition
  Transformer}. In \bibinfo{booktitle}{\emph{Proceedings of the 29th ACM
  International Conference on Multimedia}}. \bibinfo{pages}{1553--1561}.
\newblock


\bibitem[\protect\citeauthoryear{Zhao, Liu, and Wang}{Zhao
  et~al\mbox{.}}{2021}]%
        {zhao2021learning}
\bibfield{author}{\bibinfo{person}{Zengqun Zhao}, \bibinfo{person}{Qingshan
  Liu}, {and} \bibinfo{person}{Shanmin Wang}.} \bibinfo{year}{2021}\natexlab{}.
\newblock \showarticletitle{Learning deep global multi-scale and local
  attention features for facial expression recognition in the wild}.
\newblock \bibinfo{journal}{\emph{IEEE Transactions on Image Processing}}
  \bibinfo{volume}{30} (\bibinfo{year}{2021}), \bibinfo{pages}{6544--6556}.
\newblock


\end{thebibliography}


\end{document}